\documentclass[acmtog]{acmart}
\AtBeginDocument{%
  }

\setcopyright{acmlicensed}
\copyrightyear{2026}
\acmYear{2026}
\setcopyright{cc}
\setcctype{by}
\acmConference[SIGGRAPH Conference Papers '26]{Special Interest Group on Computer Graphics and Interactive Techniques Conference Conference Papers}{July 19--23, 2026}{Los Angeles, CA, USA}
\acmBooktitle{Special Interest Group on Computer Graphics and Interactive Techniques Conference Conference Papers (SIGGRAPH Conference Papers '26), July 19--23, 2026, Los Angeles, CA, USA}
\acmDOI{10.1145/3799902.3811151}
\acmISBN{979-8-4007-2554-8/2026/07}

\acmSubmissionID{875}

\usepackage[linesnumbered,ruled,vlined]{algorithm2e}
\usepackage{enumitem}

\begin{document}

%%
%% The "title" command has an optional parameter,
%% allowing the author to define a "short title" to be used in page headers.
\title{LongE2V: Long-Horizon Event-based Video Reconstruction, Prediction, and Frame Interpolation with Video Diffusion Models}

%%
%% The "author" command and its associated commands are used to define
%% the authors and their affiliations.
%% Of note is the shared affiliation of the first two authors, and the
%% "authornote" and "authornotemark" commands
%% used to denote shared contribution to the research.
\author{Cheng-De Fan}
\affiliation{%
  \institution{National Yang Ming Chiao Tung University}
  \country{Taiwan}}
\email{fansam39@gmail.com}

\author{Chun-Wei Tuan Mu}
\affiliation{%
  \institution{National Yang Ming Chiao Tung University}
  \country{Taiwan}}
\email{raytm9999@gmail.com}

\author{Chen-Wei Chang}
\affiliation{%
  \institution{National Yang Ming Chiao Tung University}
  \country{Taiwan}}
\email{steven891213.ii12@nycu.edu.tw}

\author{Chin-Yang Lin}
\affiliation{%
  \institution{National Yang Ming Chiao Tung University}
  \country{Taiwan}}
\email{linjohn0903@gmail.com}

\author{Kun-Ru Wu}
\affiliation{%
  \institution{National Yang Ming Chiao Tung University}
  \country{Taiwan}}
\email{wufish@nycu.edu.tw}

\author{Yu-Chee Tseng}
\affiliation{%
  \institution{National Yang Ming Chiao Tung University}
  \country{Taiwan}}
\email{yctseng@nycu.edu.tw}

\author{Yu-Lun Liu}
\affiliation{%
  \institution{National Yang Ming Chiao Tung University}
  \country{Taiwan}}
\email{yulunliu@cs.nycu.edu.tw}

%%
%% By default, the full list of authors will be used in the page
%% headers. Often, this list is too long, and will overlap
%% other information printed in the page headers. This command allows
%% the author to define a more concise list
%% of authors' names for this purpose.
\renewcommand{\shortauthors}{Fan et al.}

%%
%% The abstract is a short summary of the work to be presented in the
%% article.
\begin{abstract}
  Recovering high-quality video from sparse event streams is a challenging task. Regression methods often blur textures, while existing generative models struggle with long-term stability. We propose LongE2V, a novel approach that leverages pre-trained video diffusion priors to jointly handle event-based video reconstruction, prediction, and frame interpolation. By fine-tuning a foundational video model, our approach achieves high data efficiency and superior perceptual quality. We introduce Autoregressive Unrolling and Adaptive Context Switching to mitigate temporal drift in extremely long sequences. We also propose Reencoding Alignment with Cross Residual Correction to ensure precise bidirectional consistency during frame interpolation. Furthermore, Event Voxel Density Augmentation ensures robustness across varying sensor resolutions. Extensive experiments on real-world benchmarks demonstrate that LongE2V outperforms state-of-the-art methods across all three tasks, exhibiting exceptional temporal coherence and zero-shot generalization.
  Project page: \url{https://cdfan0627.github.io/LongE2V-page/}
\end{abstract}

%%
%% The code below is generated by the tool at http://dl.acm.org/ccs.cfm.
%% Please copy and paste the code instead of the example below.
%%


%%
%% Keywords. The author(s) should pick words that accurately describe
%% the work being presented. Separate the keywords with commas.
% \keywords{Event-based Vision, Video Diffusion Models, Video Reconstruction, Video Prediction, Video Frame Interpolation}
%% A "teaser" image appears between the author and affiliation
%% information and the body of the document, and typically spans the
%% page.
\begin{teaserfigure}
  \includegraphics[width=\textwidth]{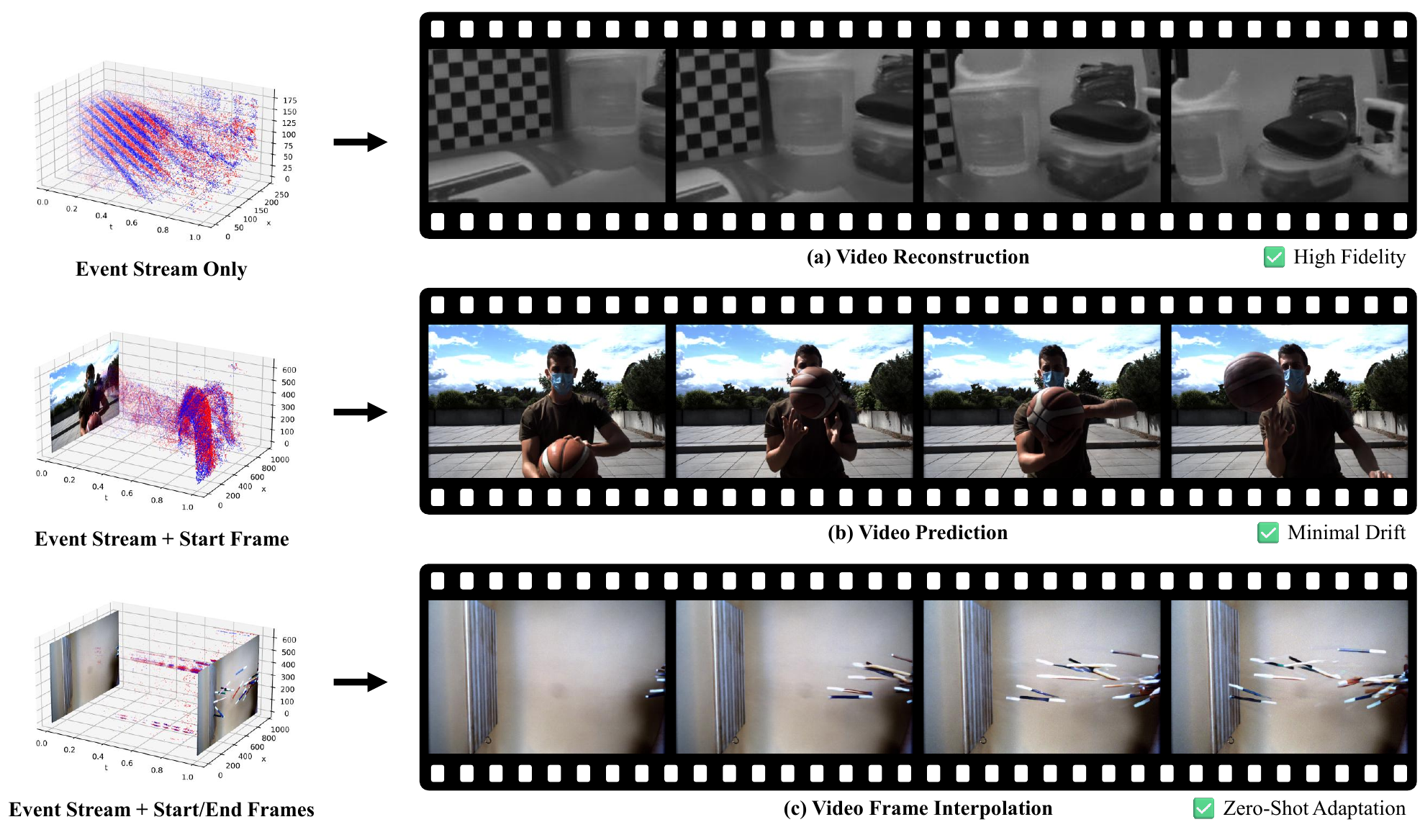}
  \caption{
  \textbf{Event-based video generation.} We leverage pre-trained video diffusion priors to address three distinct inverse problems within a single architecture. Depending on the input condition, our model performs: (a) \textbf{Video Reconstruction}, recovering high-fidelity textures from sparse event streams, (b) \textbf{Video Prediction}, generating long-term sequences from a single start frame with minimal drift via our autoregressive unrolling strategy, and (c) \textbf{Video Frame Interpolation}, achieving zero-shot adaptation to synthesize intermediate frames by leveraging event dynamics as temporal guidance. 
  % Our approach outperforms baselines in both perceptual quality and temporal stability.
  }
  \label{fig:teaser}
\end{teaserfigure}

% \received{20 February 2007}
% \received[revised]{12 March 2009}
% \received[accepted]{5 June 2009}

%%
%% This command processes the author and affiliation and title
%% information and builds the first part of the formatted document.
\maketitle

\section{Introduction}

Event cameras are bio-inspired sensors capturing asynchronous brightness changes with microsecond resolution and high dynamic range (HDR). Unlike standard cameras prone to motion blur, they excel in high-speed dynamics. However, their sparse, intensity-free output is incompatible with standard vision algorithms, making high-fidelity video recovery an inherently ill-posed problem. We address this by leveraging video diffusion models for Video Reconstruction, Prediction, and Frame Interpolation. As illustrated in Fig.~\ref{fig:teaser}, a robust solution must simultaneously recover photometric details (Reconstruction), ensure long-term coherence (Prediction), and enable zero-shot intermediate frame generation (Frame Interpolation), effectively bridging neuromorphic sensing and human-interpretable vision.

Traditional event-based video generation relies on CNNs or RNNs \cite{rebecq2019high,scheerlinck2020fast}, with pioneers like E2VID~\cite{rebecq2019high} and FireNet~\cite{scheerlinck2020fast} aggregating temporal information. Recent advances employ Transformers~\cite{weng2021event} and Hypernetworks~\cite{ercan2024hypere2vid} to enhance representation. With the rise of generative AI, Video Diffusion Models (VDMs)\cite{ho2022video,blattmann2023stable} like VDM-EVFI\cite{chen2025repurposing} have been adapted for interpolation. These methods establish strong baselines by mapping event volumes to frames and are often trained on synthetic datasets.

\begin{figure}
  \includegraphics[width=\columnwidth]{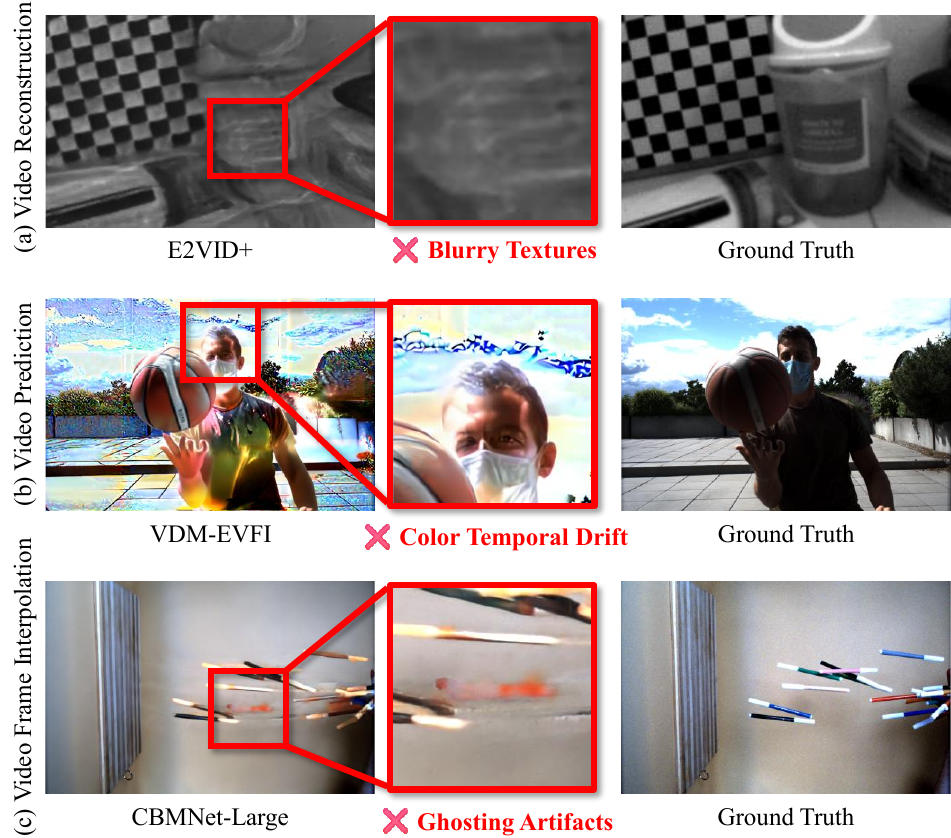}
  \caption{
  \textbf{Challenges in event-based video generation.}
  We highlight failure cases in state-of-the-art methods:
(a) Reconstruction: Regression-based methods (e.g., E2VID+\cite{stoffregen2020reducing}) suffer from ``regression-to-the-mean,'' causing \textbf{blurry textures} and detail loss.
(b) Prediction: Direct video diffusion (e.g., VDM-EVFI\cite{chen2025repurposing}) on long sequences suffers from error accumulation, leading to severe \textbf{color temporal drift}.
(c) Interpolation: Existing networks (e.g., CBMNet-Large~\cite{kim2023event}) fail to capture complex intermediate motion, producing significant \textbf{ghosting artifacts}. Our method leverages video diffusion priors to ensure high-fidelity and stable generation across all tasks.
  }
  \label{fig:motivation}
\end{figure}

Despite advancements, challenges persist in real-world scenarios. Regression-based methods like E2VID suffer from ``regression-to-the-mean,'' yielding blurry textures (Fig.\ref{fig:motivation}(a)). While diffusion models improve generation, naive application causes instability; long-term prediction suffers from severe error accumulation and drift (Fig.\ref{fig:motivation}(b)). Furthermore, interpolation methods struggle with fast, complex dynamics, producing ghosting artifacts (Fig.~\ref{fig:motivation}(c)). Crucially, many existing methods are tailored to individual tasks, often requiring separate architectures for reconstruction, prediction, and frame interpolation, thus limiting flexibility.

To address these limitations, we propose \textbf{LongE2V}, leveraging pre-trained Video Diffusion Models (CogVideoX \cite{yang2024cogvideox}) to handle reconstruction, prediction, and frame interpolation. As shown in Fig.~\ref{fig:pipeline}, we formulate these tasks as conditional generation driven by event voxels. To ensure long-term stability, we introduce Autoregressive Unrolling combined with Adaptive Context Switching, which dynamically adjusts temporal dependencies to mitigate error accumulation and drift. For interpolation, we propose Reencoding Alignment and Cross Residual Correction to resolve temporal misalignments in the 3D VAE latent space. Finally, Event Voxel Density Augmentation ensures robustness across varying sensor resolutions.

Our contributions are summarized as follows:
\begin{itemize}[leftmargin=*]
\item We propose LongE2V, leveraging pre-trained video diffusion priors to handle event-based reconstruction, prediction, and frame interpolation with superior data efficiency.
\item We introduce Autoregressive Unrolling and Adaptive Context Switching to ensure long-term stability in reconstruction and prediction, alongside Reencoding Alignment with Cross Residual Correction to ensure temporal consistency in frame interpolation.
\item We design Event Voxel Density Augmentation to achieve robust generalization across different sensor resolutions, demonstrating that our method outperforms SOTA baselines across all three tasks and achieves superior perceptual quality, stability, and zero-shot generalization.
\end{itemize}

% Prior methods for event-based video generation mainly used recurrent convolutional neural networks (RNNs) or U-Net architectures that were trained from scratch~\cite{rebecq2019high,stoffregen2020reducing,tulyakov2021time}. These approaches usually process events through voxel grids and learn to produce intensity images directly. While they work well for short-term reconstruction, they require substantial data and often fail to adapt to different sensor resolutions or complex scene dynamics without large paired datasets. Additionally, regression-based objectives can cause "regression-to-the-mean" artifacts. This results in blurry textures and a lack of sharp details. Recently, Video Diffusion Models (VDMs)~\cite{ho2022video,blattmann2023stable} have shown strong generative capabilities, but their use with event data has mainly been limited to specific tasks or faces high computational costs.

\section{Related Work}
\label{sec:related_work}
\paragraph{Event-based Video Reconstruction.}
Reconstructing intensity from relative brightness changes is ill-posed~\cite{gallego2020event}. Early optimization methods~\cite{bardow2016simultaneous,munda2018real,scheerlinck2018continuous,zhang2022formulating} were superseded by deep learning. E2VID~\cite{rebecq2019high,rebecq2019events} established recurrent U-Net baselines trained on synthetic data~\cite{rebecq2018esim}, followed by improvements in efficiency~\cite{scheerlinck2020fast,cadena2023sparse}, sim-to-real transfer~\cite{stoffregen2020reducing,cadena2021spade,ercan2024hypere2vid}, and architectures including Transformers~\cite{weng2021event}, SSL~\cite{paredes2021back,wang2024revisit}, SNNs~\cite{zhu2022event,zhu2020retina}, and GANs~\cite{wang2019event}. Recent generative approaches employ language~\cite{chen2024lase}, diffusion~\cite{liang2024e2vidiff}, or temporal residuals~\cite{zhu2024temporal}. In contrast, our method handles reconstruction, prediction, and frame interpolation via efficient fine-tuning.
\paragraph{Event-based Video Frame Interpolation.}
EVFI exploits high temporal resolution to synthesize intermediate frames. Approaches include warping-synthesis hybrids~\cite{tulyakov2021time,tulyakov2022time,kim2023event,sun2024unified,sun2023event,ma2024timelens,cho2024tta} and flow-based methods employing cycle-consistency~\cite{liu2019deep,he2022timereplayer} or adaptive computation~\cite{wu2022video,shi2023ido,liu2024event,zhang2022unifying,zhang2025evdi++}. Recently, EPA~\cite{liu2025epa} further leverages fine-grained event cues to guide hierarchical feature alignment within a perceptual space. Direct synthesis methods~\cite{paikin2021efi,liu2024video} have recently adapted pre-trained video diffusion models~\cite{chen2025repurposing}. Unlike prior work restricted to interpolation, we extend diffusion priors to reconstruction and prediction, introducing \emph{Reencoding Alignment} and \emph{Cross Residual Correction} to resolve 3D VAE temporal misalignment.
\paragraph{Video Diffusion Models.}
Diffusion models~\cite{ho2020denoising,song2020score,chao2022denoising} evolved from U-Net architectures optimizing attention and efficiency~\cite{ho2022video,ho2022imagen,singer2022make,blattmann2023align,blattmann2023stable,bar2024lumiere} to Diffusion Transformers (DiTs) achieving state-of-the-art quality~\cite{yang2024cogvideox,ma2024latte,chen2024gentron}. Foundation models now scale to minute-long generation~\cite{brooks2024video,zheng2024open,kong2024hunyuanvideo,genmo2024mochi} using flow matching~\cite{esser2024scaling,yin2025slow,chen2025goku}. Diffusion priors also transfer zero-shot to video restoration~\cite{yeh2024diffir2vr}. Building on CogVideoX~\cite{yang2024cogvideox}, we utilize events for \emph{explicit motion guidance}, unlike the implicit motion learning in text-to-video generation.
\paragraph{Controllable Video Generation.}
Control methods range from spatial conditioning via adapters or attention~\cite{zhang2023adding,mou2024t2i,chen2023control,wang2023videocomposer,guo2023animatediff,chen2026pantheon360} to fine-grained trajectory control~\cite{wang2024motionctrl,wu2024draganything,ma2024trailblazer,zhang2025tora,geng2025motion,namekata2024sg,he2024cameractrl,jin2025flovd,wang2024boximator}. Structure-guided approaches separate content from motion~\cite{esser2023structure,niu2024mofa,wang2024easycontrol,xiao2024video,chen2024narcan,huang2026generative}. We posit that event streams offer ideal structural conditions, encoding natural scene dynamics with microsecond resolution without manual specification.
\paragraph{Long-term Video Generation.}
Long video generation combats error accumulation~\cite{bengio2015scheduled,lamb2016professor} via memory or streaming architectures~\cite{henschel2025streamingt2v,shiu2025stream,yin2023nuwa,zhao2024moviedreamer} or training-free noise rescheduling~\cite{qiu2023freenoise,kim2024fifo,lu2024freelong,chen2025ouroboros,zhou2024upscale,chen2023seine}. Recent works target the train-inference gap directly~\cite{huang2025self,guo2025end,zhang2025packing,yin2025slow,yoo2023towards}. We introduce \emph{Autoregressive Unrolling} to fine-tune on predictions~\cite{bengio2015scheduled,lamb2016professor} and \emph{Adaptive Context Switch} to dynamically update context, preventing temporal drift unlike fixed-schedule methods.

\section{Method}

\subsection{Preliminaries}
\label{subsec:preliminaries}
\paragraph{Video Diffusion Models}
We adopt CogVideoX I2V~\cite{yang2024cogvideox}, which encodes video $X$ into latents $Z_0$ using a 3D VAE with $4\times$ temporal and $8\times$ spatial compression. The Diffusion Transformer (DiT) $\epsilon_\theta$ is trained to minimize the denoising objective: $\mathcal{L} = \mathbb{E}_{Z_0, t, C, \epsilon} \left[ \| \epsilon - \epsilon_\theta(Z_t, t, C) \|_2^2 \right],$
where $Z_t$ are the noisy latents and $C$ denotes conditioning signals. During inference, the model iteratively denoises random Gaussian noise to synthesize clean latents, which are decoded back to pixel space.
% We adopt CogVideoX I2V~\cite{yang2024cogvideox} as our backbone. It encodes video $X \in \mathbb{R}^{F \times H \times W \times 3}$ into latents $Z_0 \in \mathbb{R}^{\frac{F}{4} \times \frac{H}{8} \times \frac{W}{8} \times C}$ using a 3D VAE with $4\times$ temporal and $8\times$ spatial compression. Diffusion Transformer (DiT) serves as the denoiser $\epsilon_\theta$, trained to minimize the reconstruction error:
% \begin{equation} \small
% \mathcal{L} = \mathbb{E}_{Z_0, t, C, \epsilon} \left[ \| \epsilon - \epsilon_\theta(Z_t, t, C) \|_2^2 \right],
% \end{equation}
% where $Z_t$ is the noisy latent at timestep $t$, and $C$ represents conditioning signals (e.g., text prompts and reference images). During inference, the model iteratively denoises from random Gaussian noise to synthesize clean latents. These are finally decoded back to the pixel space via the 3D VAE decoder to reconstruct the final video.

\paragraph{Event Representation}
Event cameras capture asynchronous streams $\{e_i\}_{i=1}^{N}$ where $e_i = (x_i, y_i, t_i, p_i)$. To enable frame-based processing, we discretize these events into a voxel grid $V \in \mathbb{R}^{B \times H \times W}$~\cite{ercan2024hypere2vid} by accumulating polarity $p_i \in \{\pm 1\}$ into $B$ temporal bins via linear interpolation:
\begin{equation}
V(t, y, x) = \sum_{i} p_i \max(0, 1 - |t - t_i^*|) \delta(x - x_i, y - y_i),
\end{equation}
where $\delta$ is the Kronecker delta and $t_i^* \in [0, B-1]$ represents the normalized timestamp relative to duration $\Delta T$. In our experiments, we set $B = 3$.
% Event cameras capture intensity changes asynchronously as a stream $\{e_i\}_{i=1}^{N}$, where each event $e_i = (x_i, y_i, t_i, p_i)$ contains spatiotemporal coordinates and polarity $p_i \in \{+1, -1\}$. To leverage standard frame-based backbones, we discretize the sparse stream into an event voxel grid $V \in \mathbb{R}^{B \times H \times W}$ following the formulation in~\cite{zihao2018unsupervised}. Specifically, events within a duration $\Delta T$ are accumulated into $B$ temporal bins using differentiable linear interpolation:
% \begin{equation} \small
% V(t, y, x) = \sum_{i} p_i \max(0, 1 - |t - t_i^*|) \delta(x - x_i, y - y_i),
% \end{equation}
% where $\delta$ represents the Kronecker delta, and  $t_i^* = (B-1)(t_i - T_{start}) / \Delta T$ denotes the normalized timestamp mapped to $[0, B-1]$. In all our experiments, we set $B = 3$.

\subsection{Event-based Video Generation Framework}
We denote the current video frames as $X = \{x_0, \dots, x_{F-1}\}$. To align asynchronous events with discrete frames, we partition the event stream into time windows. Specifically, the event voxel $v_k$ corresponds to frame $x_k$ at timestamp $t_k$ by aggregating events within the interval $[t_{k-1}, t_k)$ via the transformation in Sec.~\ref{subsec:preliminaries}. This yields a synchronized event sequence $V = \{v_1, \dots, v_{F-1}\}$.

\paragraph{Context and Autoregressive Unrolling}
Naive long video generation often uses the last frame of the previous chunk as the initial condition for the next. However, this recursive process causes error accumulation and artifact propagation. To mitigate this, we employ context frames and context event voxels as history conditions. Specifically, context frames are defined as the continuous sequence immediately preceding the first frame of the current chunk, and context event voxels are the corresponding event voxels associated with these frames.

\begin{figure}
  \includegraphics[width=\columnwidth]{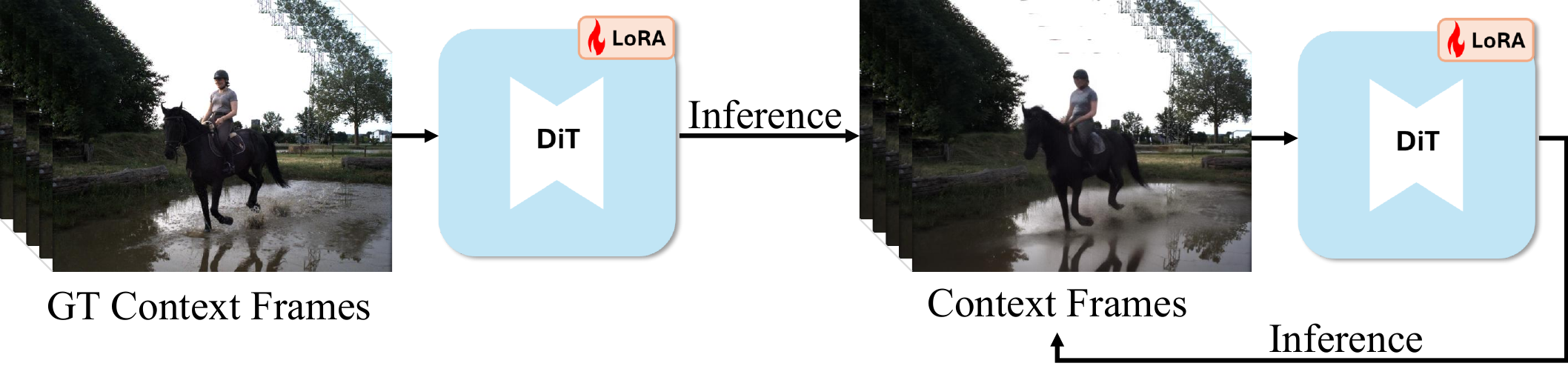}
  \caption{\textbf{Autoregressive Unrolling.} To bridge the domain gap between training and inference, we employ an iterative training strategy. Initially, the model is trained with Ground Truth (GT) context frames for convergence (left). Subsequently, we activate the unrolling mechanism by performing an inference pass to generate predictions, which then replace the GT context frames for fine-tuning (right). This iterative feedback loop forces the model to adapt to its own generation errors, mitigating error accumulation during long video generation.}
  \label{fig:autoregressive_unrolling}
\end{figure}

However, simply conditioning on history introduces a domain gap between training and inference, as the model conditions on Ground Truth context frames during training but relies on its own predictions during inference. To bridge this gap, we propose an Autoregressive Unrolling training strategy. Initially, the model is trained with GT context frames until convergence. We then activate the unrolling mechanism where predictions generated on the training set are substituted as context frames for subsequent fine-tuning, as illustrated in Fig.~\ref{fig:autoregressive_unrolling}. This iterative unrolling is repeated $T$ times, forcing the model to adapt to its own prediction errors and effectively aligning the training distribution with inference behavior.

\paragraph{Event Voxel Density Augmentation}
The spatial density of event voxels varies significantly due to diverse sensor resolutions and scene depths. To enhance robustness, we introduce a density augmentation strategy during training. Specifically, we randomly resize event voxels while preserving the aspect ratio, bounded within a dynamic range $[S_{min}, S_{max}]$. The lower bound $S_{min}$ is set slightly larger than the network input to facilitate random cropping, while $S_{max}$ is capped at $2\times$ the original resolution. Following FireNet~\cite{scheerlinck2020fast}, we normalize based on the statistics of non-zero values to preserve sparsity. Finally, a random crop is applied to match the network input resolution. To maintain spatial alignment, the identical geometric transformations are synchronously applied to the first frame, context frames, and current video frames.

\paragraph{Finetuning with Event Voxels and Context}
Following spatial alignment, we ensure temporal alignment of the input data prior to fine-tuning. Since the first frame $x_0$ is provided, we zero-pad the current event voxels $V$ at the initial timestamp to align with the subsequent frames $x_{1:F-1}$. All inputs, including $x_0$, padded $V$, and context elements, are encoded via a frozen 3D VAE. Note that constructing $V$ with $B=3$ allows the event stream to be processed natively by the VAE's standard 3-channel input. We then construct the final input latents by concatenating three temporally-aligned sequences along the channel dimension: (1) context latents with the zero-padded first-frame latent $Z_{x_0}$, (2) context latents with the noise latents $Z_t$, and (3) context event latents with the current event latents.

\begin{figure}
  \includegraphics[width=\columnwidth]{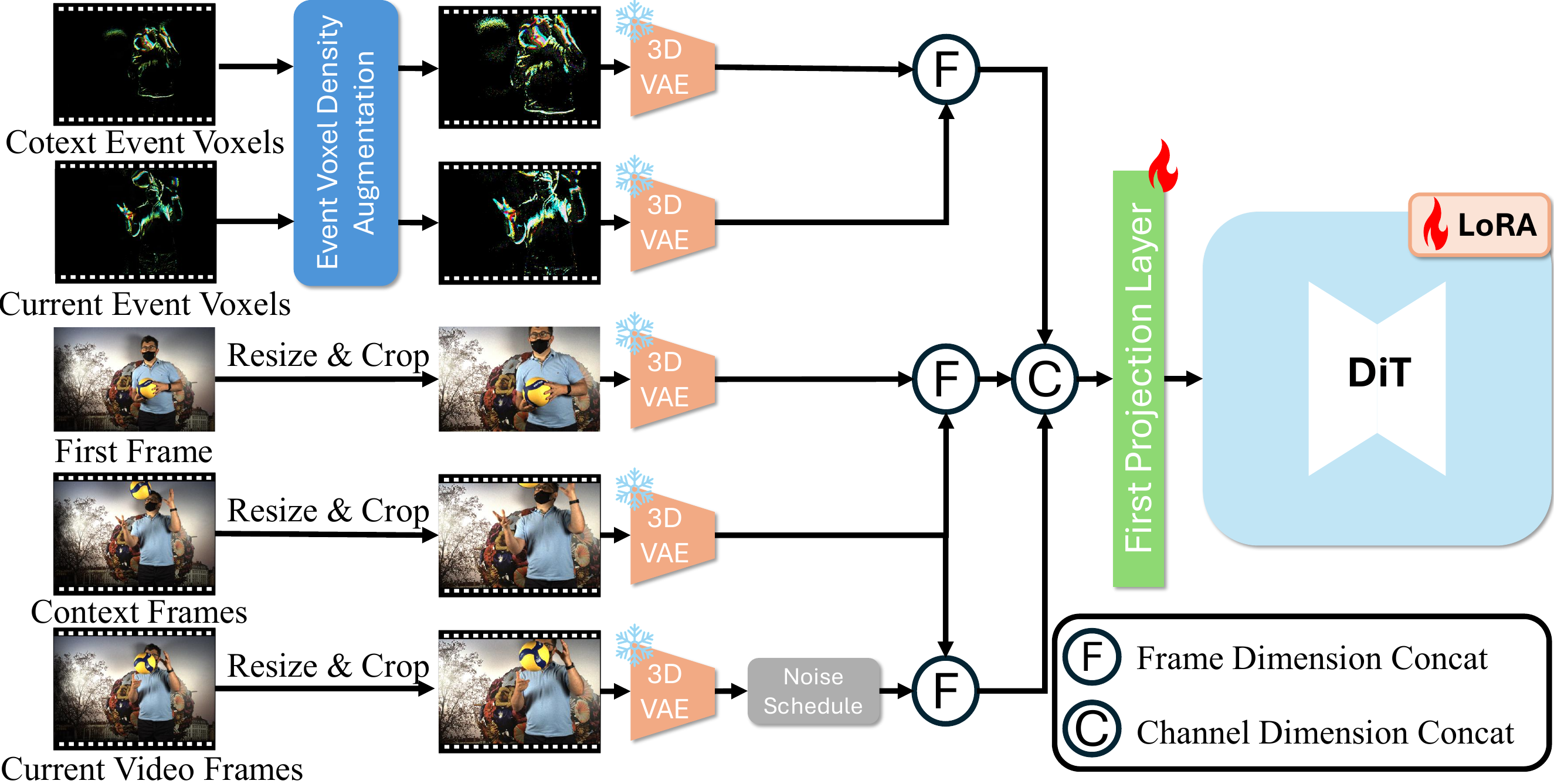}
  \caption{\textbf{Overview of our training pipeline.} To enhance robustness against sensor variations, input event voxels undergo Event Voxel Density Augmentation, and the first frame, context frames, and current video frames are synchronously resized and cropped to ensure spatial alignment. All inputs are encoded into latents via a frozen 3D VAE. These latents are aligned and fused through frame dimension concatenation and channel dimension concatenation. Finally, we expand and fully fine-tune the First Projection Layer to accommodate the additional event channels, while the DiT backbone is efficiently fine-tuned using LoRA.}
  \label{fig:pipeline}
\end{figure}

To accommodate the additional event condition, we augment the first projection layer within the patchify module, extending the weights $\mathbf{W}_{in} \in \mathbb{R}^{D \times 2C_{vae} \times K \times K}$ to $\mathbf{W}_{in}^{*} \in \mathbb{R}^{D \times 3C_{vae} \times K \times K}$. During training, we fully fine-tune this expanded layer and employ LoRA for the DiT blocks, as shown in Fig.~\ref{fig:pipeline}. The loss is calculated exclusively on the $Z_t$ component. Additionally, we apply a 5\% dropout to $Z_{x_0}$ to enhance reconstruction robustness and introduce text prompts with a 20\% probability for text-guided generation.

\subsection{Event-based Video Reconstruction, Prediction, and Frame Interpolation.}
We formulate event-based video reconstruction, prediction, and frame interpolation as tasks of event-based video generation. By adjusting the input conditions, the model adapts to specific task requirements:
\begin{itemize}[leftmargin=*]
\item \textbf{Video Reconstruction:} The model recovers photometric details solely from events. For the first chunk, start frame and context are empty, forcing the model to reconstruct the scene from scratch. (see Sec.~\ref {sec:Experiment_Settings} for details) For subsequent chunks, the last frame of the preceding chunk serves as the first frame, seamlessly transitioning the task into a prediction paradigm.
\item \textbf{Video Prediction:} Given the start frame, the event stream, and context, the model generates subsequent video frames by leveraging the start frame for initial texture guidance, the event stream for dynamic information, and the context for historical reference. For the first chunk, the context frames are populated by replicating the start frame to fill the temporal window, while context event voxels are initialized as empty (see Sec.~\ref {sec:Experiment_Settings} for details).
\item \textbf{Video Frame Interpolation:}  Given the start frame and end frame as boundary conditions, the model synthesizes intermediate frames by leveraging both forward and backward event streams. These streams act as a dense motion guide bridging the boundaries.
\end{itemize}

\paragraph{Adaptive Context Switch}
Standard autoregressive updates after every chunk cause error accumulation. To mitigate this, we introduce an Adaptive Context Switch. While the first chunk requires a mandatory update, subsequent updates are dynamic based on context relevance. We re-input the denoised latents $\hat{Z}_0$ into the DiT to extract the attention map $\mathbf{A}$ and compute the average attention weight $\mu_{attn}$ of current tokens attending to context tokens:
\begin{equation}
\mu_{attn} = \frac{1}{L \times H \times N_{curr}} \sum_{l=1}^{L} \sum_{h=1}^{H} \sum_{i \in \text{Current}} \sum_{j \in \text{Context}} \mathbf{A}^{(l, h)}_{i,j},
\label{eq:attn_score}
\end{equation}
where $L$ is the number of layers, $H$ is the number of Attention Heads, and $N_{curr}$ is the number of Current Tokens. If $\mu_{attn} \ge \tau$, the existing context is retained to prevent drift. If $\mu_{attn} < \tau$, indicating low relevance, we trigger a Context Switch with a single-attempt retry mechanism: the current generation is discarded, the context is updated to the immediate predecessor, and the chunk is regenerated. In our experiments, we set $\tau=0.05$.

\begin{figure*}[t]
  \includegraphics[width=\textwidth]{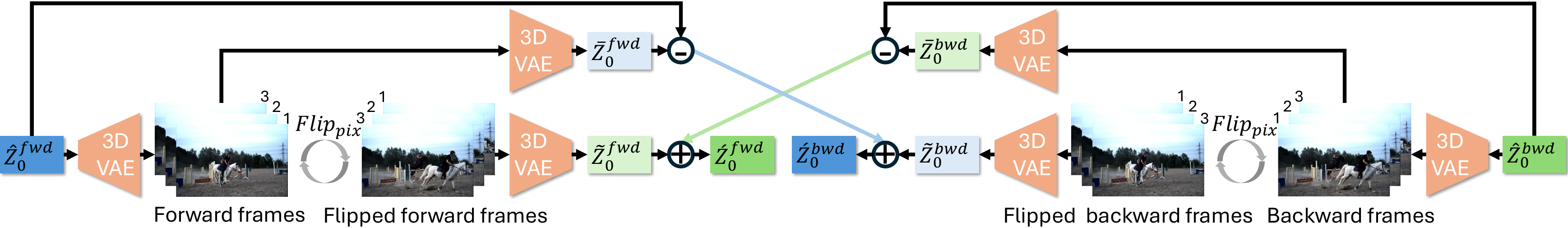}
  \caption{\textbf{Reencoding Alignment
 and Cross Residual Correction.} To address temporal misalignment caused by the discrepancy between latent-space and pixel-space flipping, we propose Reencoding Alignment. The denoised latents, $\hat{Z}_0^{fwd}$ and $\hat{Z}_0^{bwd}$, are decoded into pixel space, flipped temporally ($Flip_{pix}$), and then re-encoded via the 3D VAE to yield the aligned latents $\tilde{Z}_0^{fwd}$ and $\tilde{Z}_0^{bwd}$. To mitigate information loss inherent in this re-encoding process, we employ Cross Residual Correction. We calculate the residual difference between the original and the re-encoded latents (e.g., the subtraction node $\hat{Z}_0^{fwd} - \bar{Z}_0^{fwd}$) and inject this detail information into the opposite branch. Specifically, the forward residual is added to the backward aligned latents $\tilde{Z}_0^{bwd}$ to produce the final corrected latents $Z'^{bwd}_0$, and symmetrically, the backward residual is injected into $\tilde{Z}_0^{fwd}$ to obtain $Z'^{fwd}_0$. This symmetric Cross Injection mechanism promotes temporal consensus between branches while preserving fine-grained details. Light-colored boxes represent information loss.}
  \label{fig:vae_residual}
\end{figure*}

\paragraph{Reencoding Alignment}
We extend Time Reversal Fusion~\cite{feng2024explorative} for zero-shot interpolation. Standard methods directly flip backward latents, but due to temporal compression in 3D VAEs~\cite{yang2024cogvideox,wan2025wan}, latent and pixel space operations are not commutative:
\begin{equation} 
\text{Flip}_{lat}(Z_t) \neq \mathcal{E}(\text{Flip}_{pix}(\mathcal{D}(Z_t))),
\end{equation}
where $\mathcal{D}, \mathcal{E}$ are the decoder/encoder and $\text{Flip}_{lat/pix}$ denote temporal flipping. This discrepancy causes misalignment. We propose Reencoding Alignment to rectify this by decoding the predicted clean latents $\hat{Z}_0$, flipping in pixel space, and re-encoding:
\begin{equation} 
\tilde{Z}_0^{fwd} = \mathcal{E}(\text{Flip}_{pix}(\mathcal{D}(\hat{Z}_0^{fwd}))), \quad \tilde{Z}_0^{bwd} = \mathcal{E}(\text{Flip}_{pix}(\mathcal{D}(\hat{Z}_0^{bwd}))).
\end{equation}
This ensures precise temporal alignment between the bidirectional branches (Fig.~\ref{fig:vae_residual}).
% Our model extends to zero-shot video frame interpolation via Time Reversal Fusion \cite{feng2024explorative}, decomposing the task into forward and backward prediction processes. Standard approaches fuse these by directly flipping the backward latents $Z_t^{bwd}$ in the latent space. However, since mainstream 3D VAEs \cite{yang2024cogvideox,wan2025wan} perform temporal compression, operations in latent and pixel spaces are not commutative:
% \begin{equation} \small
% \text{Flip}_{lat}(Z_t) \neq \mathcal{E}(\text{Flip}_{pix}(\mathcal{D}(Z_t))),
% \end{equation}
% where $\mathcal{D}$ and $\mathcal{E}$ denote the decoder and encoder, while $\text{Flip}_{lat}$ and $\text{Flip}_{pix}$ represent temporal flipping in latent and pixel spaces, respectively. This discrepancy leads to temporal misalignment during fusion. To rectify this, we propose Reencoding Alignment. At each denoising step, we decode the predicted clean latents $\hat{Z}_0$ to pixel space, apply the temporal flip, and re-encode them to obtain the aligned latents:
% \begin{equation} \small
% \tilde{Z}_0^{fwd} = \mathcal{E}(\text{Flip}_{pix}(\mathcal{D}(\hat{Z}_0^{fwd}))), \quad \tilde{Z}_0^{bwd} = \mathcal{E}(\text{Flip}_{pix}(\mathcal{D}(\hat{Z}_0^{bwd}))).
% \end{equation}
% This ensures precise temporal alignment between the bidirectional branches, as shown in Fig.~\ref{fig:vae_residual}.

\paragraph{Cross Residual Correction}
Although Reencoding Alignment resolves misalignment, the extra Decode-Encode loop induces reconstruction loss. To compensate, we introduce Cross Residual Correction, inspired by LookingGlass~\cite{chang2025lookingglass}. We design a Cross Injection strategy leveraging complementary details from forward and backward branches to mutually restore information. We compute the residual between the original $\hat{Z}_0$ and re-encoded $\bar{Z}_0 = \mathcal{E}(\mathcal{D}(\hat{Z}_0))$ latents:
\begin{equation} 
\Delta \hat{Z}_0^{fwd} = \hat{Z}_0^{fwd} - \bar{Z}_0^{fwd}, \quad \Delta \hat{Z}_0^{bwd} = \hat{Z}_0^{bwd} - \bar{Z}_0^{bwd}.
\end{equation}
We then inject the opposing residual into the aligned latents:
\begin{equation} 
Z'^{bwd}_0 = \tilde{Z}_0^{bwd} + \Delta \hat{Z}_0^{fwd}, \quad Z'^{fwd}_0 = \tilde{Z}_0^{fwd} + \Delta \hat{Z}_0^{bwd}.
\end{equation}
This recovers high-frequency details and promotes Temporal Consensus (Fig.~\ref{fig:vae_residual}). Finally, the corrected latents are fused via alpha blending and re-noised for the subsequent loop.
% While Reencoding Alignment resolves misalignment, the additional Decode-Encode loop introduces reconstruction loss. To compensate, we introduce Cross Residual Correction, inspired by LookingGlass~\cite{chang2025lookingglass}, which uses residuals to preserve vibrant colors in anamorphoses. Tailored for bidirectional interpolation, we design a Cross Injection strategy that leverages the complementary details of the forward and backward branches to mutually restore information. Specifically, we define the re-encoded latent as $\bar{Z}_0 = \mathcal{E}(\mathcal{D}(\hat{Z}_0))$ and compute the residual relative to the original latent:
% \begin{equation} \small
% \Delta \hat{Z}_0^{fwd} = \hat{Z}_0^{fwd} - \bar{Z}_0^{fwd}, \quad \Delta \hat{Z}_0^{bwd} = \hat{Z}_0^{bwd} - \bar{Z}_0^{bwd}.
% \end{equation}
% Next, we inject the opposing residual into the aligned latent:
% \begin{equation} \small
% Z'^{bwd}_0 = \tilde{Z}_0^{bwd} + \Delta \hat{Z}_0^{fwd}, \quad Z'^{fwd}_0 = \tilde{Z}_0^{fwd} + \Delta \hat{Z}_0^{bwd}.
% \end{equation}
% This not only recovers high-frequency details but also promotes the bidirectional branches to reach Temporal Consensus, as illustrated in Fig.~\ref{fig:vae_residual}. Finally, the corrected $Z'^{bwd}_0$ and $Z'^{fwd}_0$ are fused with the original predictions via alpha blending and re-noised for the subsequent loop.

\section{Experiments}
\subsection{Experiment Settings}
\label{sec:Experiment_Settings}
\paragraph{Dataset.}
We train exclusively on the BS-ERGB~\cite{tulyakov2021time} training set, chosen for its high-quality real-world events and large motion, after filtering sequences with missing data.
For reconstruction and prediction, we follow the EVREAL benchmark~\cite{ercan2023evreal} on selected subsets of three real-world event datasets: ECD~\cite{mueggler2017event}, MVSEC~\cite{zhu2018multivehicle}, and HQF~\cite{stoffregen2020reducing}.
Crucially, these sequences cover a wide range of temporal durations, spanning from short clips in ECD ($\sim$300 frames) to long-term sequences in MVSEC and HQF (up to 2,740 and 2,430 frames, respectively), evaluating the model's stability over extended periods.
For frame interpolation, we evaluate on the BS-ERGB test set and HQF dataset.

\paragraph{Implementation Details.} 
We employ CogVideoX I2V as our backbone, generating 49-frame chunks at $720 \times 480$. In training, we apply LoRA ($r=64$) to the DiT blocks and full finetune the first projection layer. Training involves 3 times autoregressive unrolling. A 20-frame context is maintained during both training and inference. Specifically, for the first chunk during inference where historical context is unavailable, we apply task-specific initialization: reconstruction employs zero tensors for both the start frame latent and context latents; prediction replicates the start frame $20\times$ to populate the context; and frame interpolation replicates the start and end frames $10\times$ each to form the context. Across all tasks, the corresponding context event voxels are consistently zero-padded.
During evaluation, inspired by VDM-EVFI, we upsample inputs to mitigate detail loss from VAE encoding. Specifically, inputs below the model’s input resolution are upsampled to match it, whereas those exceeding it are upsampled by $\approx2\times$ and processed using their proposed Per-tile Denoising and Fusion strategy.

\subsection{Comparisons with SOTA Event base Reconstruction and Prediction Methods}
For reconstruction, we benchmark against SOTA E2V methods, including E2VID, FireNet, E2VID+, FireNet+, SPADE-E2VID, SSL-E2VID, ET-Net, and HyperE2VID, using pre-trained weights from EVREAL. For prediction, we compare with VDM-EVFI using its official 13-frame pre-trained weight in the forward generation setting. Since event streams lack absolute intensity information, for a fair comparison, we align the global brightness of all generated results with the ground truth sequences.

\begin{table*}[t]
    \centering
    % \small
    \caption{
    % \textbf{Quantitative reconstruction and prediction comparisons on sequences from ECD, MVSEC, and HQF datasets.}
    \textbf{Quantitative results on ECD~\cite{mueggler2017event}, MVSEC~\cite{zhu2018multivehicle}, and HQF~\cite{stoffregen2020reducing}.} Comparing our method against SOTA baselines (\textbf{\textcolor{red}{Red}}: best; \underline{\textcolor{blue}{blue}}: second). In \emph{Reconstruction}, we consistently achieve the best LPIPS scores, indicating superior perceptual quality compared to regression-based methods. In \emph{Prediction}, our method significantly outperforms VDM-EVFI across all metrics and datasets, validating our robust long-term generation capabilities.
    }
\label{tab:recon_pred}
    \resizebox{\textwidth}{!}{%
    \begin{tabular}{lccccccccccc}
    \toprule
    & & \multicolumn{3}{c}{ECD~\cite{mueggler2017event}} & \multicolumn{3}{c}{MVSEC~\cite{zhu2018multivehicle}} & \multicolumn{3}{c}{HQF~\cite{stoffregen2020reducing}} \\
    \cmidrule(lr){3-5} \cmidrule(lr){6-8} \cmidrule(lr){9-11}
    Method & Venue & PSNR $\uparrow$ & SSIM $\uparrow$ & LPIPS $\downarrow$ & PSNR $\uparrow$ & SSIM $\uparrow$ & LPIPS $\downarrow$ & PSNR $\uparrow$ & SSIM $\uparrow$ & LPIPS $\downarrow$ \\
    \midrule
    \textbf{Reconstruction} \\
    E2VID~\cite{rebecq2019high} & TPAMI 2019 &\textbf{\color{red}22.40}  &\color{blue}\underline{0.690} &0.177  &14.84  &0.354  &0.576  &15.06  &0.522  &0.362  \\
    FireNet~\cite{scheerlinck2020fast} & WACV 2020 &20.01  &0.597 &0.241  &15.51  &0.374  &0.587  &13.89  &0.445  &0.486 \\
    E2VID+~\cite{stoffregen2020reducing} & ECCV 2020 &\color{blue}\underline{22.32}  &0.676 &\color{blue}\underline{0.147}  &\color{blue}\underline{16.27}  &0.411  &0.463  &15.97  &\color{blue}\underline{0.552}  &\color{blue}\underline{0.256}  \\
    FireNet+~\cite{stoffregen2020reducing} &ECCV 2020 &21.18  &0.621 &0.215  &15.45  & 0.389 &0.498  &15.64  &0.491  &0.314  \\
    SPADE-E2VID~\cite{cadena2021spade} & TIP 2021 &19.71  &0.569 &0.301  &15.70  &0.389  &0.538  &13.36  &0.417  &0.538  \\
    SSL-E2VID~\cite{paredes2021back} & CVPR 2021 &20.00  &0.615 &0.278  &14.36  &0.353  &0.633  &13.51  &0.446  &0.474  \\
    ET-Net~\cite{weng2021event} & ICCV 2021 &22.26  &0.675 &0.159  &15.97  &0.409  &0.435  &16.17 &0.542& 0.263 \\
    HyperE2VID~\cite{ercan2024hypere2vid} & TIP 2024 &\textbf{\color{red}22.40}  &0.685 &0.157  &16.13  &\color{blue}\underline{0.417}  &\color{blue}\underline{0.434}  &\color{blue}\underline{16.42}  & 0.539 & 0.257 \\
    Ours & - &22.28  &\textbf{\color{red}0.708}&\textbf{\color{red}0.139}  &\textbf{\color{red}16.83}  &\textbf{\color{red}0.440}  &\textbf{\color{red}0.405}  &\textbf{\color{red}16.45}  &\textbf{\color{red}0.603}  &\textbf{\color{red}0.240}  \\ \midrule
    \textbf{Prediction (w/ start frame)} \\
    VDM-EVFI~\cite{chen2025repurposing} & CVPR 2025 &20.33  &0.614  &0.244  &15.46  &0.276 &0.668  &13.52  &0.373 &0.520 \\
    Ours & - &\textbf{\color{red}24.40}  &\textbf{\color{red}0.771}  &\textbf{\color{red}0.110}  &\textbf{\color{red}18.18}  &\textbf{\color{red}0.502}  &\textbf{\color{red}0.359}  &\textbf{\color{red}16.67}  &\textbf{\color{red}0.619}  &\textbf{\color{red}0.229}  \\
\bottomrule
\end{tabular}
}
\end{table*}

\begin{figure*}
  \includegraphics[width=\textwidth]{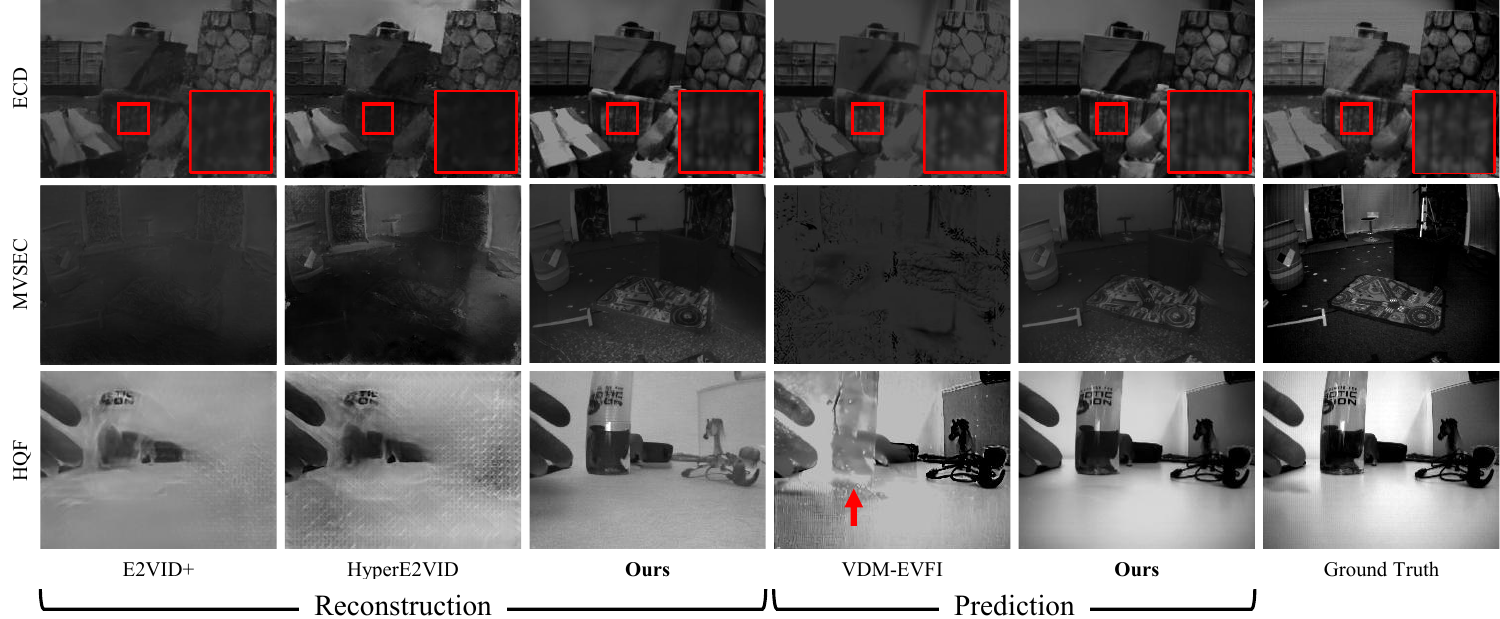}
  \caption{
  \textbf{Qualitative comparisons on ECD~\cite{mueggler2017event}, MVSEC~\cite{zhu2018multivehicle}, and HQF~\cite{stoffregen2020reducing} datasets.}
  Our LongE2V recovers high-frequency textures where regression baselines (E2VID+, HyperE2VID) suffer from blurring (Row 1). In prediction tasks, we avoid the severe noise accumulation and ghosting artifacts (red arrow) seen in VDM-EVFI (Rows 2–3), maintaining superior structural fidelity and temporal stability.
  % ~\yulunliu{Put citation for each dataset and method.}
  % \textbf{Quantitative reconstruction and prediction comparisons on sequences from ECD (rows 1-2), MVSEC (rows 3-4), HQF (rows 5-7).}
  }
  \label{fig:recon_pred}
\end{figure*}

As shown in Fig.~\ref{fig:recon_pred}, the 2th row illustrates the later stages of the sequences, where existing SOTA models suffer from severe error accumulation in both tasks. The 3th row depicts the early stages; here, SOTA E2V models struggle to reconstruct high-quality images due to the initialization limitations of their recurrent architectures. Notably, VDM-EVFI exhibits noticeable error accumulation even in the early part of the sequence. Both Tab.~\ref{tab:recon_pred} and Fig.~\ref{fig:recon_pred} demonstrate that our method consistently outperforms baselines across all three datasets, validating its superiority in handling both short and long-term sequences. More visual comparisons are shown in Fig.~\ref{fig:appendix_recon}.

\subsection{Comparisons with SOTA Event base Video Frame Interpolation Methods}
Although our model was trained exclusively for reconstruction and prediction, we extended it to Event-based Video Frame Interpolation (EVFI) without any fine-tuning. By utilizing the identical weights, we evaluated our method in a zero-shot setting against dedicated SOTA EVFI baselines (CBMNet-Large~\cite{kim2023event} and TLXNet+~\cite{ma2024timelens}) on interpolating 31 frames.

\begin{figure}
  \includegraphics[width=\columnwidth]{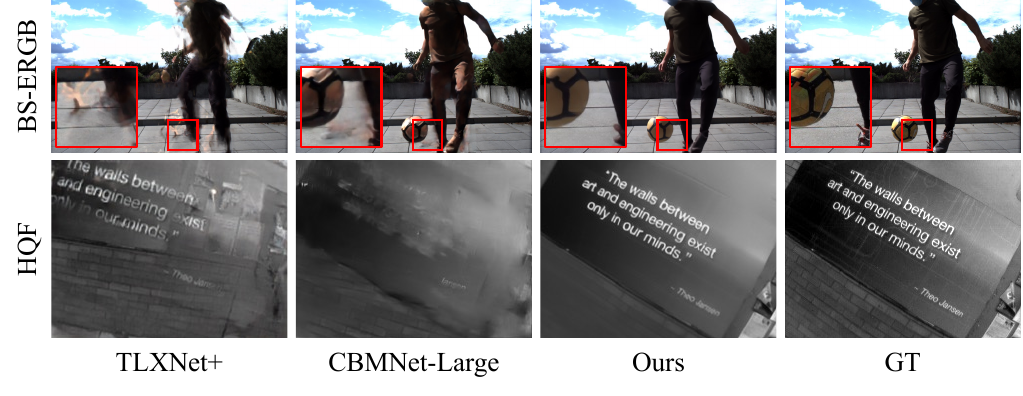}
  \caption{
  \textbf{Zero-shot interpolation on BS-ERGB and HQF.} Baselines (TLXNet+, CBMNet-Large) suffer from structural collapse or blur under large motion (\emph{Top}), whereas our LongE2V captures accurate dynamics. On fine text (\emph{Bottom}), our Reencoding Alignment eliminates the ghosting seen in baselines, ensuring legibility.
  % \textbf{Quantitative frame interpolation comparisons on sequences from BS-ERGB (rows 1-2), HQF (rows 3-4).
  }
  \label{fig:interpolation}
\end{figure}

\begin{table}[t]
    \centering
    % \small
    \caption{
   \textbf{Interpolation (31 skips) on ERGB~\cite{tulyakov2021time} and HQF~\cite{stoffregen2020reducing}.} Comparing our \emph{zero-shot} method vs. \emph{supervised} baselines (\textcolor{red}{\textbf{Red}}: best; \textcolor{blue}{\underline{blue}}: second), we excel in \emph{LPIPS} and generalization. Unlike regression baselines, which blur textures to boost PSNR, our generative approach preserves high-frequency details.
    % \textbf{Quantitative interpolation results (31 skips) of existing methods and our proposed method on sequences from BS-ERGB~\cite{tulyakov2021time}, and HQF~\cite{stoffregen2020reducing} datasets.}
    }
\label{tab:interpolation}
    \resizebox{\columnwidth}{!}{%
\begin{tabular}{lccccccc}
\toprule
& Training & \multicolumn{3}{c}{BS-ERGB} & \multicolumn{3}{c}{HQF} \\ \cmidrule(lr){3-5} \cmidrule(lr){6-8}
 % &  \multicolumn{3}{c|}{31skips}  & \multicolumn{3}{c}{31skips} \\ \cline{2-7} 
 % ~\cite{kim2023event}~\cite{ma2024timelens}
Method & setting & PSNR $\uparrow$ & SSIM $\uparrow$ & LPIPS $\downarrow$ & PSNR $\uparrow$ & SSIM $\uparrow$ & LPIPS $\downarrow$  \\ \midrule
CBMNet-Large & Supervised &\color{red}\textbf{24.59}  &\color{red}\textbf{0.767}  &0.170  &\color{blue}\underline{24.55} &\color{blue}\underline{0.767} &\color{blue}\underline{0.166} \\
TLXNet+ & Supervised & 18.85  &0.641  &0.226  &17.18 &0.484 &0.352  \\
Ours & Zero-shot &\color{blue}\underline{24.40}  &\color{blue}\underline{0.744}  &\color{red}\textbf{0.124}  &\color{red}\textbf{25.39}  &\color{red}\textbf{0.800} &\color{red}\textbf{0.105} \\ \bottomrule
\end{tabular}
}
\end{table}

Visual comparisons reveal significant limitations in the baselines when handling large-frame interpolation on real-world data. As shown in Row 1 of Fig.~\ref{fig:interpolation}, CBMNet-Large suffers from severe ghosting and color artifacts in dynamic human motions, while TLXNet+ fails to preserve structural integrity. Furthermore, in Rows 2, our method is the only one capable of reconstructing clear, readable text, whereas CBMNet-Large results in severe blur, and TLXNet+ produces misaligned, ghosted outputs. Both Tab.~\ref{tab:interpolation} and Fig.~\ref{fig:interpolation} confirm that our zero-shot approach surpasses existing SOTA methods designed specifically for EVFI. More visual comparisons are shown in Fig.~\ref{fig:appendix_interpolation}.

% We intended to include EPA~\cite{liu2025epa} and VDM-EVFI in this comparison. However, EPA could not be evaluated due to the lack of public pretrained weights, full training code, and training data. VDM-EVFI was excluded because its underlying architecture (based on SVD) restricts the number of generated frames, making it incapable of performing the 31-frame interpolation task.

\subsection{Ablation Study}

\begin{figure}
  \includegraphics[width=\columnwidth]{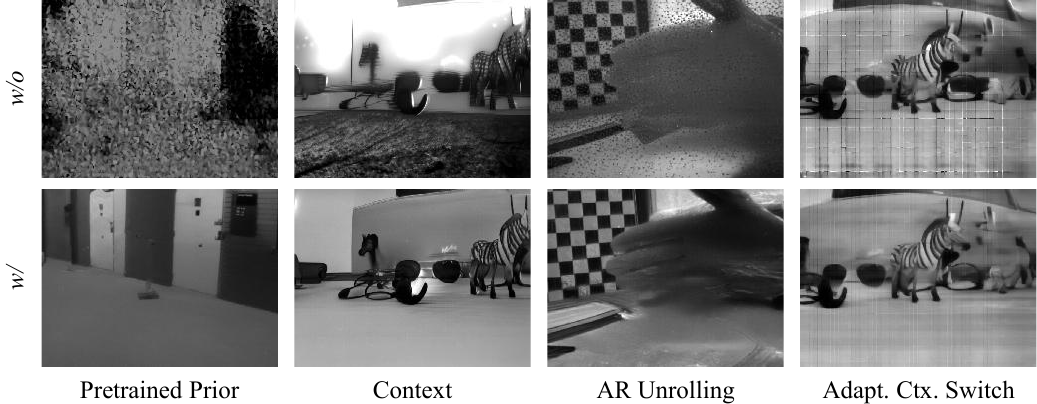}
  \caption{\textbf{Qualitative ablation on reconstruction.} \emph{Top}: Ablated variants; \emph{Bottom}: Full Method. w/o Pretrained prior yields noise; w/o Context causes structural ambiguity; w/o Autoregressive unrolling leads to point artifacts (drift); and w/o Adaptive context switch causes grid artifacts. Our method ensures stable, high-fidelity results.
  }
  \label{fig:ablation_recon}
\end{figure}

\paragraph{Ablation Study on Event-based Video Reconstruction}
We analyze component effectiveness in Fig. \ref{fig:ablation_recon} and Tab. \ref{tab:ablation_recon}. First, the pretrained prior is critical (Row 1); training the diffusion backbone from scratch on limited data fails to converge, yielding pure noise. Second, relying solely on event voxels without context mechanisms causes severe error accumulation and artifacts (Row 2). Third, removing Autoregressive Unrolling and Adaptive Context Switch (Row 3) exposes a training-inference domain gap; training with Ground Truth context while inferring with predictions causes drift, evidenced by point artifacts. Fourth, ablating only the Adaptive Context Switch (Row 4) by forcing updates after every chunk results in error accumulation, as shown by grid artifacts. These results confirm that each component is vital for minimizing error accumulation and maintaining stability.

\begin{table}[t]
\centering
% \small
\caption{
\textbf{Ablation of reconstruction components on HQF dataset.} 
The \emph{pretrained prior} is essential for convergence (Row 1 vs. 2). Including \emph{context}, \emph{Autoregressive unrolling}, \emph{Adaptive Context Switch} effectively mitigates long-term drift (Rows 3, 4, and 5), allowing the full method to achieve the best performance across all metrics.
}
\label{tab:ablation_recon}
    \resizebox{\columnwidth}{!}{%
\begin{tabular}{cccc|ccc}
    \toprule
    Pretrained & & AR & Adaptive \\
    prior &  Context &  unroll. & ctx. switch  & PSNR↑ & SSIM↑ & LPIPS↓ \\
    \midrule
    & \checkmark  &\checkmark &\checkmark &10.26  &0.094 &0.746  \\
    \checkmark & &  & &13.81  &0.508  &0.320  \\
    \checkmark & \checkmark & & &15.63 &0.561  &0.278 \\
    \checkmark & \checkmark & \checkmark  &  &16.42  &0.595  &0.247  \\
    \checkmark & \checkmark & \checkmark  &\checkmark  &\textbf{16.45}  &\textbf{0.603}  &\textbf{0.240}  \\
    \bottomrule
\end{tabular}
}
\end{table}

\begin{figure}
  \includegraphics[width=\columnwidth]{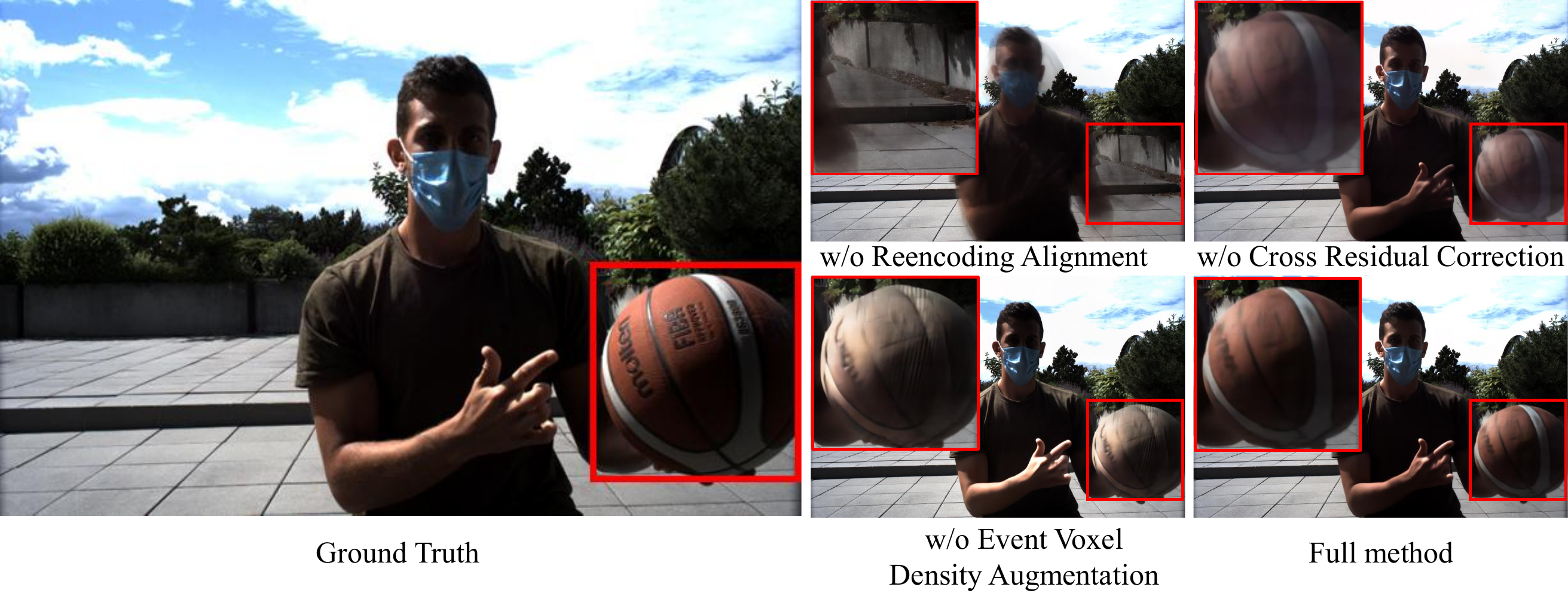}
  \caption{
  \textbf{Visual ablation on interpolation.} \emph{w/o Reencoding Alignment} causes ghosting due to latents misalignment; \emph{w/o Cross Residual Correction} blurs fine details due to VAE loss; and \emph{w/o Event Voxel Density Augmentation} yields artifacts from density mismatch. Our \emph{Full Method} restores sharp, coherent details comparable to \emph{Ground Truth}.
  }
  \label{fig:ablation_inter}
\end{figure}

\paragraph{Ablation Study on Event-based Video Frame Interpolation}
We validate our method by ablating Reencoding Alignment, Cross Residual Correction, and Event Voxel Data Augmentation (Tab. \ref{tab:ablation_inter}, Fig. \ref{fig:ablation_inter}). Without Reencoding Alignment, directly flipping latents causes temporal misalignment, resulting in significantly blurred dynamic figures. This confirms the necessity of our decode-flip-encode process for spatial coherence. Removing Cross Residual Correction leads to ghosting and semi-transparent artifacts due to 3D VAE information loss; restoring this via cross-injection is crucial for Temporal Consensus, improving LPIPS by 0.037. Finally, omitting Event Voxel Data Augmentation exposes the model to a density mismatch between training and inference, where upsampling the input resolution at test time dilutes event density per voxel, leading to unstable generation with color deviations and physical artifacts such as black stripes on the basketball.

\begin{table}[t]
\centering
\small
\caption{
\textbf{Ablation of interpolation components on BS-ERGB dataset.} Reencoding Alignment is critical for structural fidelity, while Cross Residual Correction enhances perceptual quality (LPIPS). Event Voxel Density Augmentation improves robustness, with the full method achieving superior results.
}
\label{tab:ablation_inter}
    % \resizebox{\columnwidth}{!}{%
\begin{tabular}{ccc|ccc}
    \toprule
    Reencoding & Cross res. & Event voxel \\
    alignment & correction & density aug. & PSNR↑ & SSIM↑ & LPIPS↓ \\
    \midrule
    & &\checkmark  &20.58 &0.670  &0.202   \\
    \checkmark& &\checkmark    &24.14 &0.734 &0.161  \\
    \checkmark&\checkmark & &22.58  &0.728  &0.129  \\
    \checkmark&\checkmark & \checkmark   &\textbf{24.40}  &\textbf{0.744}  &\textbf{0.124}  \\
    \bottomrule
\end{tabular}
% }
\end{table}

\subsection{Text-Guided Event Video Colorization}
\begin{figure}
  \includegraphics[width=\columnwidth]{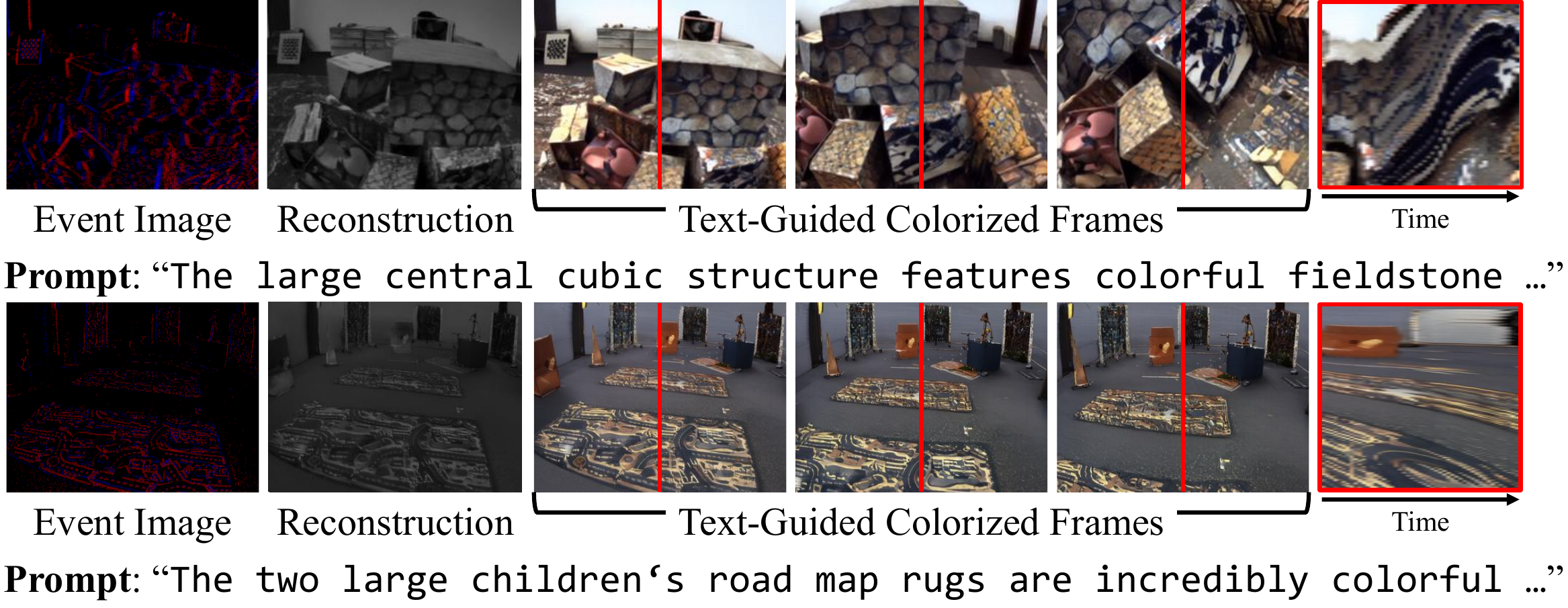}
  \caption{
\textbf{Text-Guided Event Video Colorization.} \emph{Left to Right}: Input events; standard reconstruction (geometry baseline); text-stylized frames; and a temporal XT-slice (at \textbf{\textcolor{red}{red line}}). The results demonstrate our model effectively decouples event-driven motion from text-defined appearance, while the XT-slice confirms the generated textures are temporally coherent.
}
\label{fig:prompt_edit}
\end{figure}

Leveraging the inherent text-guided generation capabilities of our backbone, CogVideoX I2V, we incorporate text prompts during training with a 20\% probability. This strategy enables the model to adapt to a multi-modal conditioning setting, learning to synthesize videos based on both text prompts and event voxels. As demonstrated in Fig.~\ref{fig:prompt_edit}, when provided with only event voxels and a text prompt, the model successfully reconstructs the structural textures of the scene from the event data while applying coloration and stylization based on the textual description, effectively achieving event video colorization.

\section{Conclusion}
\label{sec:conclusion}

We presented LongE2V, leveraging video diffusion priors for event-based reconstruction, prediction, and frame interpolation. We mitigate temporal drift via Autoregressive Unrolling and Adaptive Context Switching, while ensuring interpolation consistency using Reencoding Alignment and Cross Residual Correction. Experiments confirm LongE2V outperforms state-of-the-art methods in perceptual quality and robustness. Future work will focus on accelerating inference and exploring efficient memory mechanisms for long-term consistency.
% In this work, we presented UniE2V, a unified framework leveraging pre-trained video diffusion priors for event-based reconstruction, interpolation, and prediction. By introducing Autoregressive Unrolling and Adaptive Context Switching, we effectively mitigate temporal drift in long sequences. Additionally, our Reencoding Alignment and Cross Residual Correction ensure precise bidirectional consistency for interpolation. Extensive experiments confirm UniE2V outperforms state-of-the-art methods in perceptual quality and robustness. Future work will focus on accelerating inference latency and exploring efficient memory mechanisms to further enhance long-term temporal consistency.

\begin{acks}
This research was funded by the National Science and Technology Council, Taiwan, under Grants NSTC 112-2222-E-A49-004-MY2 and 113-2628-E-A49-023-. The authors are grateful to Google, NVIDIA, and MediaTek Inc. for their generous donations. Yu-Lun Liu acknowledges the Yushan Young Fellow Program by the MOE in Taiwan.
\end{acks}

%%
%% The next two lines define the bibliography style to be used, and
%% the bibliography file.
\bibliographystyle{ACM-Reference-Format}
\bibliography{sample-base}

%%% -*-BibTeX-*-
%%% Do NOT edit. File created by BibTeX with style
%%% ACM-Reference-Format-Journals [18-Jan-2012].

\begin{thebibliography}{103}

%%% ====================================================================
%%% NOTE TO THE USER: you can override these defaults by providing
%%% customized versions of any of these macros before the \bibliography
%%% command.  Each of them MUST provide its own final punctuation,
%%% except for \shownote{} and \showURL{}.  The latter two
%%% do not use final punctuation, in order to avoid confusing it with
%%% the Web address.
%%%
%%% To suppress output of a particular field, define its macro to expand
%%% to an empty string, or better, \unskip, like this:
%%%
%%% \newcommand{\showURL}[1]{\unskip}   % LaTeX syntax
%%%
%%% \def \showURL #1{\unskip}           % plain TeX syntax
%%%
%%% ====================================================================

\ifx \showCODEN    \undefined \def \showCODEN     #1{\unskip}     \fi
\ifx \showISBNx    \undefined \def \showISBNx     #1{\unskip}     \fi
\ifx \showISBNxiii \undefined \def \showISBNxiii  #1{\unskip}     \fi
\ifx \showISSN     \undefined \def \showISSN      #1{\unskip}     \fi
\ifx \showLCCN     \undefined \def \showLCCN      #1{\unskip}     \fi
\ifx \shownote     \undefined \def \shownote      #1{#1}          \fi
\ifx \showarticletitle \undefined \def \showarticletitle #1{#1}   \fi
\ifx \showURL      \undefined \def \showURL       {\relax}        \fi
% The following commands are used for tagged output and should be
% invisible to TeX
\providecommand\bibfield[2]{#2}
\providecommand\bibinfo[2]{#2}
\providecommand\natexlab[1]{#1}
\providecommand\showeprint[2][]{arXiv:#2}

\bibitem[Bar-Tal et~al\mbox{.}(2024)]%
        {bar2024lumiere}
\bibfield{author}{\bibinfo{person}{Omer Bar-Tal}, \bibinfo{person}{Hila Chefer}, \bibinfo{person}{Omer Tov}, \bibinfo{person}{Charles Herrmann}, \bibinfo{person}{Roni Paiss}, \bibinfo{person}{Shiran Zada}, \bibinfo{person}{Ariel Ephrat}, \bibinfo{person}{Junhwa Hur}, \bibinfo{person}{Guanghui Liu}, \bibinfo{person}{Amit Raj}, {et~al\mbox{.}}} \bibinfo{year}{2024}\natexlab{}.
\newblock \showarticletitle{Lumiere: A space-time diffusion model for video generation}. In \bibinfo{booktitle}{\emph{SIGGRAPH Asia 2024 Conference Papers}}. \bibinfo{pages}{1--11}.
\newblock


\bibitem[Bardow et~al\mbox{.}(2016)]%
        {bardow2016simultaneous}
\bibfield{author}{\bibinfo{person}{Patrick Bardow}, \bibinfo{person}{Andrew~J Davison}, {and} \bibinfo{person}{Stefan Leutenegger}.} \bibinfo{year}{2016}\natexlab{}.
\newblock \showarticletitle{Simultaneous optical flow and intensity estimation from an event camera}. In \bibinfo{booktitle}{\emph{Proceedings of the IEEE conference on computer vision and pattern recognition}}. \bibinfo{pages}{884--892}.
\newblock


\bibitem[Bengio et~al\mbox{.}(2015)]%
        {bengio2015scheduled}
\bibfield{author}{\bibinfo{person}{Samy Bengio}, \bibinfo{person}{Oriol Vinyals}, \bibinfo{person}{Navdeep Jaitly}, {and} \bibinfo{person}{Noam Shazeer}.} \bibinfo{year}{2015}\natexlab{}.
\newblock \showarticletitle{Scheduled sampling for sequence prediction with recurrent neural networks}.
\newblock \bibinfo{journal}{\emph{Advances in neural information processing systems}}  \bibinfo{volume}{28} (\bibinfo{year}{2015}).
\newblock


\bibitem[Blattmann et~al\mbox{.}(2023a)]%
        {blattmann2023stable}
\bibfield{author}{\bibinfo{person}{Andreas Blattmann}, \bibinfo{person}{Tim Dockhorn}, \bibinfo{person}{Sumith Kulal}, \bibinfo{person}{Daniel Mendelevitch}, \bibinfo{person}{Maciej Kilian}, \bibinfo{person}{Dominik Lorenz}, \bibinfo{person}{Yam Levi}, \bibinfo{person}{Zion English}, \bibinfo{person}{Vikram Voleti}, \bibinfo{person}{Adam Letts}, {et~al\mbox{.}}} \bibinfo{year}{2023}\natexlab{a}.
\newblock \showarticletitle{Stable video diffusion: Scaling latent video diffusion models to large datasets}.
\newblock \bibinfo{journal}{\emph{arXiv preprint arXiv:2311.15127}} (\bibinfo{year}{2023}).
\newblock


\bibitem[Blattmann et~al\mbox{.}(2023b)]%
        {blattmann2023align}
\bibfield{author}{\bibinfo{person}{Andreas Blattmann}, \bibinfo{person}{Robin Rombach}, \bibinfo{person}{Huan Ling}, \bibinfo{person}{Tim Dockhorn}, \bibinfo{person}{Seung~Wook Kim}, \bibinfo{person}{Sanja Fidler}, {and} \bibinfo{person}{Karsten Kreis}.} \bibinfo{year}{2023}\natexlab{b}.
\newblock \showarticletitle{Align your latents: High-resolution video synthesis with latent diffusion models}. In \bibinfo{booktitle}{\emph{Proceedings of the IEEE/CVF conference on computer vision and pattern recognition}}. \bibinfo{pages}{22563--22575}.
\newblock


\bibitem[Brooks et~al\mbox{.}(2024)]%
        {brooks2024video}
\bibfield{author}{\bibinfo{person}{Tim Brooks}, \bibinfo{person}{Bill Peebles}, \bibinfo{person}{Connor Holmes}, \bibinfo{person}{Will DePue}, \bibinfo{person}{Yufei Guo}, \bibinfo{person}{Li Jing}, \bibinfo{person}{David Schnurr}, \bibinfo{person}{Joe Taylor}, \bibinfo{person}{Troy Luhman}, \bibinfo{person}{Eric Luhman}, {et~al\mbox{.}}} \bibinfo{year}{2024}\natexlab{}.
\newblock \showarticletitle{Video generation models as world simulators}.
\newblock \bibinfo{journal}{\emph{OpenAI Blog}} \bibinfo{volume}{1}, \bibinfo{number}{8} (\bibinfo{year}{2024}), \bibinfo{pages}{1}.
\newblock


\bibitem[Cadena et~al\mbox{.}(2021)]%
        {cadena2021spade}
\bibfield{author}{\bibinfo{person}{Pablo Rodrigo~Gantier Cadena}, \bibinfo{person}{Yeqiang Qian}, \bibinfo{person}{Chunxiang Wang}, {and} \bibinfo{person}{Ming Yang}.} \bibinfo{year}{2021}\natexlab{}.
\newblock \showarticletitle{Spade-e2vid: Spatially-adaptive denormalization for event-based video reconstruction}.
\newblock \bibinfo{journal}{\emph{IEEE Transactions on Image Processing}}  \bibinfo{volume}{30} (\bibinfo{year}{2021}), \bibinfo{pages}{2488--2500}.
\newblock


\bibitem[Cadena et~al\mbox{.}(2023)]%
        {cadena2023sparse}
\bibfield{author}{\bibinfo{person}{Pablo Rodrigo~Gantier Cadena}, \bibinfo{person}{Yeqiang Qian}, \bibinfo{person}{Chunxiang Wang}, {and} \bibinfo{person}{Ming Yang}.} \bibinfo{year}{2023}\natexlab{}.
\newblock \showarticletitle{Sparse-e2vid: A sparse convolutional model for event-based video reconstruction trained with real event noise}. In \bibinfo{booktitle}{\emph{Proceedings of the IEEE/CVF Conference on Computer Vision and Pattern Recognition}}. \bibinfo{pages}{4150--4158}.
\newblock


\bibitem[Chang et~al\mbox{.}(2025)]%
        {chang2025lookingglass}
\bibfield{author}{\bibinfo{person}{Pascal Chang}, \bibinfo{person}{Sergio Sancho}, \bibinfo{person}{Jingwei Tang}, \bibinfo{person}{Markus Gross}, {and} \bibinfo{person}{Vinicius Azevedo}.} \bibinfo{year}{2025}\natexlab{}.
\newblock \showarticletitle{LookingGlass: Generative Anamorphoses via Laplacian Pyramid Warping}. In \bibinfo{booktitle}{\emph{Proceedings of the Computer Vision and Pattern Recognition Conference}}. \bibinfo{pages}{24--33}.
\newblock


\bibitem[Chao et~al\mbox{.}(2022)]%
        {chao2022denoising}
\bibfield{author}{\bibinfo{person}{Chen-Hao Chao}, \bibinfo{person}{Wei-Fang Sun}, \bibinfo{person}{Bo-Wun Cheng}, \bibinfo{person}{Yi-Chen Lo}, \bibinfo{person}{Chia-Che Chang}, \bibinfo{person}{Yu-Lun Liu}, \bibinfo{person}{Yu-Lin Chang}, \bibinfo{person}{Chia-Ping Chen}, {and} \bibinfo{person}{Chun-Yi Lee}.} \bibinfo{year}{2022}\natexlab{}.
\newblock \showarticletitle{Denoising likelihood score matching for conditional score-based data generation}.
\newblock \bibinfo{journal}{\emph{arXiv preprint arXiv:2203.14206}} (\bibinfo{year}{2022}).
\newblock


\bibitem[Chen et~al\mbox{.}(2025a)]%
        {chen2025repurposing}
\bibfield{author}{\bibinfo{person}{Jingxi Chen}, \bibinfo{person}{Brandon~Y Feng}, \bibinfo{person}{Haoming Cai}, \bibinfo{person}{Tianfu Wang}, \bibinfo{person}{Levi Burner}, \bibinfo{person}{Dehao Yuan}, \bibinfo{person}{Cornelia Fermuller}, \bibinfo{person}{Christopher~A Metzler}, {and} \bibinfo{person}{Yiannis Aloimonos}.} \bibinfo{year}{2025}\natexlab{a}.
\newblock \showarticletitle{Repurposing pre-trained video diffusion models for event-based video interpolation}. In \bibinfo{booktitle}{\emph{Proceedings of the Computer Vision and Pattern Recognition Conference}}. \bibinfo{pages}{12456--12466}.
\newblock


\bibitem[Chen et~al\mbox{.}(2025c)]%
        {chen2025ouroboros}
\bibfield{author}{\bibinfo{person}{Jingyuan Chen}, \bibinfo{person}{Fuchen Long}, \bibinfo{person}{Jie An}, \bibinfo{person}{Zhaofan Qiu}, \bibinfo{person}{Ting Yao}, \bibinfo{person}{Jiebo Luo}, {and} \bibinfo{person}{Tao Mei}.} \bibinfo{year}{2025}\natexlab{c}.
\newblock \showarticletitle{Ouroboros-diffusion: Exploring consistent content generation in tuning-free long video diffusion}. In \bibinfo{booktitle}{\emph{Proceedings of the AAAI Conference on Artificial Intelligence}}, Vol.~\bibinfo{volume}{39}. \bibinfo{pages}{2079--2087}.
\newblock


\bibitem[Chen et~al\mbox{.}(2024b)]%
        {chen2024lase}
\bibfield{author}{\bibinfo{person}{Kanghao Chen}, \bibinfo{person}{Hangyu Li}, \bibinfo{person}{Jiazhou Zhou}, \bibinfo{person}{Zeyu Wang}, {and} \bibinfo{person}{Lin Wang}.} \bibinfo{year}{2024}\natexlab{b}.
\newblock \showarticletitle{Lase-e2v: Towards language-guided semantic-aware event-to-video reconstruction}.
\newblock \bibinfo{journal}{\emph{Advances in Neural Information Processing Systems}}  \bibinfo{volume}{37} (\bibinfo{year}{2024}), \bibinfo{pages}{70406--70430}.
\newblock


\bibitem[Chen et~al\mbox{.}(2025b)]%
        {chen2025goku}
\bibfield{author}{\bibinfo{person}{Shoufa Chen}, \bibinfo{person}{Chongjian Ge}, \bibinfo{person}{Yuqi Zhang}, \bibinfo{person}{Yida Zhang}, \bibinfo{person}{Fengda Zhu}, \bibinfo{person}{Hao Yang}, \bibinfo{person}{Hongxiang Hao}, \bibinfo{person}{Hui Wu}, \bibinfo{person}{Zhichao Lai}, \bibinfo{person}{Yifei Hu}, {et~al\mbox{.}}} \bibinfo{year}{2025}\natexlab{b}.
\newblock \showarticletitle{Goku: Flow based video generative foundation models}. In \bibinfo{booktitle}{\emph{Proceedings of the Computer Vision and Pattern Recognition Conference}}. \bibinfo{pages}{23516--23527}.
\newblock


\bibitem[Chen et~al\mbox{.}(2024c)]%
        {chen2024gentron}
\bibfield{author}{\bibinfo{person}{Shoufa Chen}, \bibinfo{person}{Mengmeng Xu}, \bibinfo{person}{Jiawei Ren}, \bibinfo{person}{Yuren Cong}, \bibinfo{person}{Sen He}, \bibinfo{person}{Yanping Xie}, \bibinfo{person}{Animesh Sinha}, \bibinfo{person}{Ping Luo}, \bibinfo{person}{Tao Xiang}, {and} \bibinfo{person}{Juan-Manuel Perez-Rua}.} \bibinfo{year}{2024}\natexlab{c}.
\newblock \showarticletitle{Gentron: Diffusion transformers for image and video generation}. In \bibinfo{booktitle}{\emph{Proceedings of the IEEE/CVF Conference on Computer Vision and Pattern Recognition}}. \bibinfo{pages}{6441--6451}.
\newblock


\bibitem[Chen et~al\mbox{.}(2024a)]%
        {chen2024narcan}
\bibfield{author}{\bibinfo{person}{Ting-Hsuan Chen}, \bibinfo{person}{Jiewen Chan}, \bibinfo{person}{Hau-Shiang Shiu}, \bibinfo{person}{Shih-Han Yen}, \bibinfo{person}{Chang-Han Yeh}, {and} \bibinfo{person}{Yu-Lun Liu}.} \bibinfo{year}{2024}\natexlab{a}.
\newblock \showarticletitle{Narcan: Natural refined canonical image with integration of diffusion prior for video editing}.
\newblock \bibinfo{journal}{\emph{Advances in Neural Information Processing Systems}}  \bibinfo{volume}{37} (\bibinfo{year}{2024}), \bibinfo{pages}{36097--36120}.
\newblock


\bibitem[Chen et~al\mbox{.}(2026)]%
        {chen2026pantheon360}
\bibfield{author}{\bibinfo{person}{Ting-Hsuan Chen}, \bibinfo{person}{Ying-Huan Chen}, \bibinfo{person}{Tao Tu}, \bibinfo{person}{Jie-Ying Lee}, \bibinfo{person}{Cho-Ying Wu}, \bibinfo{person}{Fangzhou Lin}, \bibinfo{person}{Hengyuan Zhang}, \bibinfo{person}{David Paz}, \bibinfo{person}{Xinyu Huang}, \bibinfo{person}{Yuliang Guo}, {et~al\mbox{.}}} \bibinfo{year}{2026}\natexlab{}.
\newblock \showarticletitle{Pantheon360: Taming Digital Twin Generation via 3D-Aware 360deg Video Diffusion}. In \bibinfo{booktitle}{\emph{Proceedings of the IEEE/CVF Conference on Computer Vision and Pattern Recognition}}. \bibinfo{pages}{11138--11149}.
\newblock


\bibitem[Chen et~al\mbox{.}(2023a)]%
        {chen2023control}
\bibfield{author}{\bibinfo{person}{Weifeng Chen}, \bibinfo{person}{Yatai Ji}, \bibinfo{person}{Jie Wu}, \bibinfo{person}{Hefeng Wu}, \bibinfo{person}{Pan Xie}, \bibinfo{person}{Jiashi Li}, \bibinfo{person}{Xin Xia}, \bibinfo{person}{Xuefeng Xiao}, {and} \bibinfo{person}{Liang Lin}.} \bibinfo{year}{2023}\natexlab{a}.
\newblock \showarticletitle{Control-a-video: Controllable text-to-video generation with diffusion models}.
\newblock \bibinfo{journal}{\emph{arXiv e-prints}} (\bibinfo{year}{2023}), \bibinfo{pages}{arXiv--2305}.
\newblock


\bibitem[Chen et~al\mbox{.}(2023b)]%
        {chen2023seine}
\bibfield{author}{\bibinfo{person}{Xinyuan Chen}, \bibinfo{person}{Yaohui Wang}, \bibinfo{person}{Lingjun Zhang}, \bibinfo{person}{Shaobin Zhuang}, \bibinfo{person}{Xin Ma}, \bibinfo{person}{Jiashuo Yu}, \bibinfo{person}{Yali Wang}, \bibinfo{person}{Dahua Lin}, \bibinfo{person}{Yu Qiao}, {and} \bibinfo{person}{Ziwei Liu}.} \bibinfo{year}{2023}\natexlab{b}.
\newblock \showarticletitle{Seine: Short-to-long video diffusion model for generative transition and prediction}. In \bibinfo{booktitle}{\emph{The Twelfth International Conference on Learning Representations}}.
\newblock


\bibitem[Cho et~al\mbox{.}(2024)]%
        {cho2024tta}
\bibfield{author}{\bibinfo{person}{Hoonhee Cho}, \bibinfo{person}{Taewoo Kim}, \bibinfo{person}{Yuhwan Jeong}, {and} \bibinfo{person}{Kuk-Jin Yoon}.} \bibinfo{year}{2024}\natexlab{}.
\newblock \showarticletitle{TTA-EVF: test-time adaptation for event-based video frame interpolation via reliable pixel and sample estimation}. In \bibinfo{booktitle}{\emph{Proceedings of the IEEE/CVF Conference on Computer Vision and Pattern Recognition}}. \bibinfo{pages}{25701--25711}.
\newblock


\bibitem[Ercan et~al\mbox{.}(2023)]%
        {ercan2023evreal}
\bibfield{author}{\bibinfo{person}{Burak Ercan}, \bibinfo{person}{Onur Eker}, \bibinfo{person}{Aykut Erdem}, {and} \bibinfo{person}{Erkut Erdem}.} \bibinfo{year}{2023}\natexlab{}.
\newblock \showarticletitle{Evreal: Towards a comprehensive benchmark and analysis suite for event-based video reconstruction}. In \bibinfo{booktitle}{\emph{Proceedings of the IEEE/CVF Conference on Computer Vision and Pattern Recognition}}. \bibinfo{pages}{3943--3952}.
\newblock


\bibitem[Ercan et~al\mbox{.}(2024)]%
        {ercan2024hypere2vid}
\bibfield{author}{\bibinfo{person}{Burak Ercan}, \bibinfo{person}{Onur Eker}, \bibinfo{person}{Canberk Saglam}, \bibinfo{person}{Aykut Erdem}, {and} \bibinfo{person}{Erkut Erdem}.} \bibinfo{year}{2024}\natexlab{}.
\newblock \showarticletitle{Hypere2vid: Improving event-based video reconstruction via hypernetworks}.
\newblock \bibinfo{journal}{\emph{IEEE Transactions on Image Processing}}  \bibinfo{volume}{33} (\bibinfo{year}{2024}), \bibinfo{pages}{1826--1837}.
\newblock


\bibitem[Esser et~al\mbox{.}(2023)]%
        {esser2023structure}
\bibfield{author}{\bibinfo{person}{Patrick Esser}, \bibinfo{person}{Johnathan Chiu}, \bibinfo{person}{Parmida Atighehchian}, \bibinfo{person}{Jonathan Granskog}, {and} \bibinfo{person}{Anastasis Germanidis}.} \bibinfo{year}{2023}\natexlab{}.
\newblock \showarticletitle{Structure and content-guided video synthesis with diffusion models}. In \bibinfo{booktitle}{\emph{Proceedings of the IEEE/CVF international conference on computer vision}}. \bibinfo{pages}{7346--7356}.
\newblock


\bibitem[Esser et~al\mbox{.}(2024)]%
        {esser2024scaling}
\bibfield{author}{\bibinfo{person}{Patrick Esser}, \bibinfo{person}{Sumith Kulal}, \bibinfo{person}{Andreas Blattmann}, \bibinfo{person}{Rahim Entezari}, \bibinfo{person}{Jonas M{\"u}ller}, \bibinfo{person}{Harry Saini}, \bibinfo{person}{Yam Levi}, \bibinfo{person}{Dominik Lorenz}, \bibinfo{person}{Axel Sauer}, \bibinfo{person}{Frederic Boesel}, {et~al\mbox{.}}} \bibinfo{year}{2024}\natexlab{}.
\newblock \showarticletitle{Scaling rectified flow transformers for high-resolution image synthesis}. In \bibinfo{booktitle}{\emph{Forty-first international conference on machine learning}}.
\newblock


\bibitem[Feng et~al\mbox{.}(2024)]%
        {feng2024explorative}
\bibfield{author}{\bibinfo{person}{Haiwen Feng}, \bibinfo{person}{Zheng Ding}, \bibinfo{person}{Zhihao Xia}, \bibinfo{person}{Simon Niklaus}, \bibinfo{person}{Victoria Abrevaya}, \bibinfo{person}{Michael~J Black}, {and} \bibinfo{person}{Xuaner Zhang}.} \bibinfo{year}{2024}\natexlab{}.
\newblock \showarticletitle{Explorative inbetweening of time and space}. In \bibinfo{booktitle}{\emph{European Conference on Computer Vision}}. Springer, \bibinfo{pages}{378--395}.
\newblock


\bibitem[Gallego et~al\mbox{.}(2020)]%
        {gallego2020event}
\bibfield{author}{\bibinfo{person}{Guillermo Gallego}, \bibinfo{person}{Tobi Delbr{\"u}ck}, \bibinfo{person}{Garrick Orchard}, \bibinfo{person}{Chiara Bartolozzi}, \bibinfo{person}{Brian Taba}, \bibinfo{person}{Andrea Censi}, \bibinfo{person}{Stefan Leutenegger}, \bibinfo{person}{Andrew~J Davison}, \bibinfo{person}{J{\"o}rg Conradt}, \bibinfo{person}{Kostas Daniilidis}, {et~al\mbox{.}}} \bibinfo{year}{2020}\natexlab{}.
\newblock \showarticletitle{Event-based vision: A survey}.
\newblock \bibinfo{journal}{\emph{IEEE transactions on pattern analysis and machine intelligence}} \bibinfo{volume}{44}, \bibinfo{number}{1} (\bibinfo{year}{2020}), \bibinfo{pages}{154--180}.
\newblock


\bibitem[Geng et~al\mbox{.}(2025)]%
        {geng2025motion}
\bibfield{author}{\bibinfo{person}{Daniel Geng}, \bibinfo{person}{Charles Herrmann}, \bibinfo{person}{Junhwa Hur}, \bibinfo{person}{Forrester Cole}, \bibinfo{person}{Serena Zhang}, \bibinfo{person}{Tobias Pfaff}, \bibinfo{person}{Tatiana Lopez-Guevara}, \bibinfo{person}{Yusuf Aytar}, \bibinfo{person}{Michael Rubinstein}, \bibinfo{person}{Chen Sun}, {et~al\mbox{.}}} \bibinfo{year}{2025}\natexlab{}.
\newblock \showarticletitle{Motion prompting: Controlling video generation with motion trajectories}. In \bibinfo{booktitle}{\emph{Proceedings of the Computer Vision and Pattern Recognition Conference}}. \bibinfo{pages}{1--12}.
\newblock


\bibitem[Guo et~al\mbox{.}(2025)]%
        {guo2025end}
\bibfield{author}{\bibinfo{person}{Yuwei Guo}, \bibinfo{person}{Ceyuan Yang}, \bibinfo{person}{Hao He}, \bibinfo{person}{Yang Zhao}, \bibinfo{person}{Meng Wei}, \bibinfo{person}{Zhenheng Yang}, \bibinfo{person}{Weilin Huang}, {and} \bibinfo{person}{Dahua Lin}.} \bibinfo{year}{2025}\natexlab{}.
\newblock \showarticletitle{End-to-End Training for Autoregressive Video Diffusion via Self-Resampling}.
\newblock \bibinfo{journal}{\emph{arXiv preprint arXiv:2512.15702}} (\bibinfo{year}{2025}).
\newblock


\bibitem[Guo et~al\mbox{.}(2023)]%
        {guo2023animatediff}
\bibfield{author}{\bibinfo{person}{Yuwei Guo}, \bibinfo{person}{Ceyuan Yang}, \bibinfo{person}{Anyi Rao}, \bibinfo{person}{Zhengyang Liang}, \bibinfo{person}{Yaohui Wang}, \bibinfo{person}{Yu Qiao}, \bibinfo{person}{Maneesh Agrawala}, \bibinfo{person}{Dahua Lin}, {and} \bibinfo{person}{Bo Dai}.} \bibinfo{year}{2023}\natexlab{}.
\newblock \showarticletitle{Animatediff: Animate your personalized text-to-image diffusion models without specific tuning}.
\newblock \bibinfo{journal}{\emph{arXiv preprint arXiv:2307.04725}} (\bibinfo{year}{2023}).
\newblock


\bibitem[He et~al\mbox{.}(2024)]%
        {he2024cameractrl}
\bibfield{author}{\bibinfo{person}{Hao He}, \bibinfo{person}{Yinghao Xu}, \bibinfo{person}{Yuwei Guo}, \bibinfo{person}{Gordon Wetzstein}, \bibinfo{person}{Bo Dai}, \bibinfo{person}{Hongsheng Li}, {and} \bibinfo{person}{Ceyuan Yang}.} \bibinfo{year}{2024}\natexlab{}.
\newblock \showarticletitle{Cameractrl: Enabling camera control for text-to-video generation}.
\newblock \bibinfo{journal}{\emph{arXiv preprint arXiv:2404.02101}} (\bibinfo{year}{2024}).
\newblock


\bibitem[He et~al\mbox{.}(2022)]%
        {he2022timereplayer}
\bibfield{author}{\bibinfo{person}{Weihua He}, \bibinfo{person}{Kaichao You}, \bibinfo{person}{Zhendong Qiao}, \bibinfo{person}{Xu Jia}, \bibinfo{person}{Ziyang Zhang}, \bibinfo{person}{Wenhui Wang}, \bibinfo{person}{Huchuan Lu}, \bibinfo{person}{Yaoyuan Wang}, {and} \bibinfo{person}{Jianxing Liao}.} \bibinfo{year}{2022}\natexlab{}.
\newblock \showarticletitle{Timereplayer: Unlocking the potential of event cameras for video interpolation}. In \bibinfo{booktitle}{\emph{Proceedings of the IEEE/CVF Conference on Computer Vision and Pattern Recognition}}. \bibinfo{pages}{17804--17813}.
\newblock


\bibitem[Henschel et~al\mbox{.}(2025)]%
        {henschel2025streamingt2v}
\bibfield{author}{\bibinfo{person}{Roberto Henschel}, \bibinfo{person}{Levon Khachatryan}, \bibinfo{person}{Hayk Poghosyan}, \bibinfo{person}{Daniil Hayrapetyan}, \bibinfo{person}{Vahram Tadevosyan}, \bibinfo{person}{Zhangyang Wang}, \bibinfo{person}{Shant Navasardyan}, {and} \bibinfo{person}{Humphrey Shi}.} \bibinfo{year}{2025}\natexlab{}.
\newblock \showarticletitle{Streamingt2v: Consistent, dynamic, and extendable long video generation from text}. In \bibinfo{booktitle}{\emph{Proceedings of the Computer Vision and Pattern Recognition Conference}}. \bibinfo{pages}{2568--2577}.
\newblock


\bibitem[Ho et~al\mbox{.}(2022a)]%
        {ho2022imagen}
\bibfield{author}{\bibinfo{person}{Jonathan Ho}, \bibinfo{person}{William Chan}, \bibinfo{person}{Chitwan Saharia}, \bibinfo{person}{Jay Whang}, \bibinfo{person}{Ruiqi Gao}, \bibinfo{person}{Alexey Gritsenko}, \bibinfo{person}{Diederik~P Kingma}, \bibinfo{person}{Ben Poole}, \bibinfo{person}{Mohammad Norouzi}, \bibinfo{person}{David~J Fleet}, {et~al\mbox{.}}} \bibinfo{year}{2022}\natexlab{a}.
\newblock \showarticletitle{Imagen video: High definition video generation with diffusion models}.
\newblock \bibinfo{journal}{\emph{arXiv preprint arXiv:2210.02303}} (\bibinfo{year}{2022}).
\newblock


\bibitem[Ho et~al\mbox{.}(2020)]%
        {ho2020denoising}
\bibfield{author}{\bibinfo{person}{Jonathan Ho}, \bibinfo{person}{Ajay Jain}, {and} \bibinfo{person}{Pieter Abbeel}.} \bibinfo{year}{2020}\natexlab{}.
\newblock \showarticletitle{Denoising diffusion probabilistic models}.
\newblock \bibinfo{journal}{\emph{Advances in neural information processing systems}}  \bibinfo{volume}{33} (\bibinfo{year}{2020}), \bibinfo{pages}{6840--6851}.
\newblock


\bibitem[Ho et~al\mbox{.}(2022b)]%
        {ho2022video}
\bibfield{author}{\bibinfo{person}{Jonathan Ho}, \bibinfo{person}{Tim Salimans}, \bibinfo{person}{Alexey Gritsenko}, \bibinfo{person}{William Chan}, \bibinfo{person}{Mohammad Norouzi}, {and} \bibinfo{person}{David~J Fleet}.} \bibinfo{year}{2022}\natexlab{b}.
\newblock \showarticletitle{Video diffusion models}.
\newblock \bibinfo{journal}{\emph{Advances in neural information processing systems}}  \bibinfo{volume}{35} (\bibinfo{year}{2022}), \bibinfo{pages}{8633--8646}.
\newblock


\bibitem[Huang et~al\mbox{.}(2025)]%
        {huang2025self}
\bibfield{author}{\bibinfo{person}{Xun Huang}, \bibinfo{person}{Zhengqi Li}, \bibinfo{person}{Guande He}, \bibinfo{person}{Mingyuan Zhou}, {and} \bibinfo{person}{Eli Shechtman}.} \bibinfo{year}{2025}\natexlab{}.
\newblock \showarticletitle{Self Forcing: Bridging the Train-Test Gap in Autoregressive Video Diffusion}.
\newblock \bibinfo{journal}{\emph{arXiv preprint arXiv:2506.08009}} (\bibinfo{year}{2025}).
\newblock


\bibitem[Huang et~al\mbox{.}(2026)]%
        {huang2026generative}
\bibfield{author}{\bibinfo{person}{Zheng-Hui Huang}, \bibinfo{person}{Zhixiang Wang}, \bibinfo{person}{Jiaming Tan}, \bibinfo{person}{Ruihan Yu}, \bibinfo{person}{Yidan Zhang}, \bibinfo{person}{Bo Zheng}, \bibinfo{person}{Yu-Lun Liu}, \bibinfo{person}{Yung-Yu Chuang}, {and} \bibinfo{person}{Kaipeng Zhang}.} \bibinfo{year}{2026}\natexlab{}.
\newblock \showarticletitle{Generative World Renderer}.
\newblock \bibinfo{journal}{\emph{arXiv preprint arXiv:2604.02329}} (\bibinfo{year}{2026}).
\newblock


\bibitem[Jin et~al\mbox{.}(2025)]%
        {jin2025flovd}
\bibfield{author}{\bibinfo{person}{Wonjoon Jin}, \bibinfo{person}{Qi Dai}, \bibinfo{person}{Chong Luo}, \bibinfo{person}{Seung-Hwan Baek}, {and} \bibinfo{person}{Sunghyun Cho}.} \bibinfo{year}{2025}\natexlab{}.
\newblock \showarticletitle{Flovd: Optical flow meets video diffusion model for enhanced camera-controlled video synthesis}. In \bibinfo{booktitle}{\emph{Proceedings of the Computer Vision and Pattern Recognition Conference}}. \bibinfo{pages}{2040--2049}.
\newblock


\bibitem[Kim et~al\mbox{.}(2024)]%
        {kim2024fifo}
\bibfield{author}{\bibinfo{person}{Jihwan Kim}, \bibinfo{person}{Junoh Kang}, \bibinfo{person}{Jinyoung Choi}, {and} \bibinfo{person}{Bohyung Han}.} \bibinfo{year}{2024}\natexlab{}.
\newblock \showarticletitle{Fifo-diffusion: Generating infinite videos from text without training}.
\newblock \bibinfo{journal}{\emph{Advances in Neural Information Processing Systems}}  \bibinfo{volume}{37} (\bibinfo{year}{2024}), \bibinfo{pages}{89834--89868}.
\newblock


\bibitem[Kim et~al\mbox{.}(2023)]%
        {kim2023event}
\bibfield{author}{\bibinfo{person}{Taewoo Kim}, \bibinfo{person}{Yujeong Chae}, \bibinfo{person}{Hyun-Kurl Jang}, {and} \bibinfo{person}{Kuk-Jin Yoon}.} \bibinfo{year}{2023}\natexlab{}.
\newblock \showarticletitle{Event-based video frame interpolation with cross-modal asymmetric bidirectional motion fields}. In \bibinfo{booktitle}{\emph{Proceedings of the IEEE/CVF Conference on Computer Vision and Pattern Recognition}}. \bibinfo{pages}{18032--18042}.
\newblock


\bibitem[Kong et~al\mbox{.}(2024)]%
        {kong2024hunyuanvideo}
\bibfield{author}{\bibinfo{person}{Weijie Kong}, \bibinfo{person}{Qi Tian}, \bibinfo{person}{Zijian Zhang}, \bibinfo{person}{Rox Min}, \bibinfo{person}{Zuozhuo Dai}, \bibinfo{person}{Jin Zhou}, \bibinfo{person}{Jiangfeng Xiong}, \bibinfo{person}{Xin Li}, \bibinfo{person}{Bo Wu}, \bibinfo{person}{Jianwei Zhang}, {et~al\mbox{.}}} \bibinfo{year}{2024}\natexlab{}.
\newblock \showarticletitle{Hunyuanvideo: A systematic framework for large video generative models}.
\newblock \bibinfo{journal}{\emph{arXiv preprint arXiv:2412.03603}} (\bibinfo{year}{2024}).
\newblock


\bibitem[Lamb et~al\mbox{.}(2016)]%
        {lamb2016professor}
\bibfield{author}{\bibinfo{person}{Alex~M Lamb}, \bibinfo{person}{Anirudh~Goyal ALIAS PARTH~GOYAL}, \bibinfo{person}{Ying Zhang}, \bibinfo{person}{Saizheng Zhang}, \bibinfo{person}{Aaron~C Courville}, {and} \bibinfo{person}{Yoshua Bengio}.} \bibinfo{year}{2016}\natexlab{}.
\newblock \showarticletitle{Professor forcing: A new algorithm for training recurrent networks}.
\newblock \bibinfo{journal}{\emph{Advances in neural information processing systems}}  \bibinfo{volume}{29} (\bibinfo{year}{2016}).
\newblock


\bibitem[Liang et~al\mbox{.}(2024)]%
        {liang2024e2vidiff}
\bibfield{author}{\bibinfo{person}{Jinxiu Liang}, \bibinfo{person}{Bohan Yu}, \bibinfo{person}{Yixin Yang}, \bibinfo{person}{Yiming Han}, {and} \bibinfo{person}{Boxin Shi}.} \bibinfo{year}{2024}\natexlab{}.
\newblock \showarticletitle{E2VIDiff: Perceptual Events-to-Video Reconstruction using Diffusion Priors}.
\newblock \bibinfo{journal}{\emph{arXiv preprint arXiv:2407.08231}} (\bibinfo{year}{2024}).
\newblock


\bibitem[Liu et~al\mbox{.}(2024b)]%
        {liu2024event}
\bibfield{author}{\bibinfo{person}{Yuhan Liu}, \bibinfo{person}{Yongjian Deng}, \bibinfo{person}{Hao Chen}, \bibinfo{person}{Bochen Xie}, \bibinfo{person}{Youfu Li}, {and} \bibinfo{person}{Zhen Yang}.} \bibinfo{year}{2024}\natexlab{b}.
\newblock \showarticletitle{Event-based video frame interpolation with edge guided motion refinement}.
\newblock \bibinfo{journal}{\emph{arXiv preprint arXiv:2404.18156}} (\bibinfo{year}{2024}).
\newblock


\bibitem[Liu et~al\mbox{.}(2024a)]%
        {liu2024video}
\bibfield{author}{\bibinfo{person}{Yuhan Liu}, \bibinfo{person}{Yongjian Deng}, \bibinfo{person}{Hao Chen}, {and} \bibinfo{person}{Zhen Yang}.} \bibinfo{year}{2024}\natexlab{a}.
\newblock \showarticletitle{Video frame interpolation via direct synthesis with the event-based reference}. In \bibinfo{booktitle}{\emph{Proceedings of the IEEE/CVF Conference on Computer Vision and Pattern Recognition}}. \bibinfo{pages}{8477--8487}.
\newblock


\bibitem[Liu et~al\mbox{.}(2025)]%
        {liu2025epa}
\bibfield{author}{\bibinfo{person}{Yuhan Liu}, \bibinfo{person}{LingHui Fu}, \bibinfo{person}{Zhen Yang}, \bibinfo{person}{Hao Chen}, \bibinfo{person}{Youfu Li}, {and} \bibinfo{person}{Yongjian Deng}.} \bibinfo{year}{2025}\natexlab{}.
\newblock \showarticletitle{{EPA}: Boosting Event-based Video Frame Interpolation with Perceptually Aligned Learning}. In \bibinfo{booktitle}{\emph{The Thirty-ninth Annual Conference on Neural Information Processing Systems}}.
\newblock
\urldef\tempurl%
\url{https://openreview.net/forum?id=zxZPpVoCNO}
\showURL{%
\tempurl}


\bibitem[Liu et~al\mbox{.}(2019)]%
        {liu2019deep}
\bibfield{author}{\bibinfo{person}{Yu-Lun Liu}, \bibinfo{person}{Yi-Tung Liao}, \bibinfo{person}{Yen-Yu Lin}, {and} \bibinfo{person}{Yung-Yu Chuang}.} \bibinfo{year}{2019}\natexlab{}.
\newblock \showarticletitle{Deep video frame interpolation using cyclic frame generation}. In \bibinfo{booktitle}{\emph{Proceedings of the AAAI Conference on Artificial Intelligence}}, Vol.~\bibinfo{volume}{33}. \bibinfo{pages}{8794--8802}.
\newblock


\bibitem[Lu et~al\mbox{.}(2024)]%
        {lu2024freelong}
\bibfield{author}{\bibinfo{person}{Yu Lu}, \bibinfo{person}{Yuanzhi Liang}, \bibinfo{person}{Linchao Zhu}, {and} \bibinfo{person}{Yi Yang}.} \bibinfo{year}{2024}\natexlab{}.
\newblock \showarticletitle{Freelong: Training-free long video generation with spectralblend temporal attention}.
\newblock \bibinfo{journal}{\emph{Advances in Neural Information Processing Systems}}  \bibinfo{volume}{37} (\bibinfo{year}{2024}), \bibinfo{pages}{131434--131455}.
\newblock


\bibitem[Ma et~al\mbox{.}(2024b)]%
        {ma2024trailblazer}
\bibfield{author}{\bibinfo{person}{Wan-Duo~Kurt Ma}, \bibinfo{person}{John~P Lewis}, {and} \bibinfo{person}{W~Bastiaan Kleijn}.} \bibinfo{year}{2024}\natexlab{b}.
\newblock \showarticletitle{Trailblazer: Trajectory control for diffusion-based video generation}. In \bibinfo{booktitle}{\emph{SIGGRAPH Asia 2024 Conference Papers}}. \bibinfo{pages}{1--11}.
\newblock


\bibitem[Ma et~al\mbox{.}(2024c)]%
        {ma2024latte}
\bibfield{author}{\bibinfo{person}{Xin Ma}, \bibinfo{person}{Yaohui Wang}, \bibinfo{person}{Xinyuan Chen}, \bibinfo{person}{Gengyun Jia}, \bibinfo{person}{Ziwei Liu}, \bibinfo{person}{Yuan-Fang Li}, \bibinfo{person}{Cunjian Chen}, {and} \bibinfo{person}{Yu Qiao}.} \bibinfo{year}{2024}\natexlab{c}.
\newblock \showarticletitle{Latte: Latent diffusion transformer for video generation}.
\newblock \bibinfo{journal}{\emph{arXiv preprint arXiv:2401.03048}} (\bibinfo{year}{2024}).
\newblock


\bibitem[Ma et~al\mbox{.}(2024a)]%
        {ma2024timelens}
\bibfield{author}{\bibinfo{person}{Yongrui Ma}, \bibinfo{person}{Shi Guo}, \bibinfo{person}{Yutian Chen}, \bibinfo{person}{Tianfan Xue}, {and} \bibinfo{person}{Jinwei Gu}.} \bibinfo{year}{2024}\natexlab{a}.
\newblock \showarticletitle{Timelens-xl: Real-time event-based video frame interpolation with large motion}. In \bibinfo{booktitle}{\emph{European Conference on Computer Vision}}. Springer, \bibinfo{pages}{178--194}.
\newblock


\bibitem[Mou et~al\mbox{.}(2024)]%
        {mou2024t2i}
\bibfield{author}{\bibinfo{person}{Chong Mou}, \bibinfo{person}{Xintao Wang}, \bibinfo{person}{Liangbin Xie}, \bibinfo{person}{Yanze Wu}, \bibinfo{person}{Jian Zhang}, \bibinfo{person}{Zhongang Qi}, {and} \bibinfo{person}{Ying Shan}.} \bibinfo{year}{2024}\natexlab{}.
\newblock \showarticletitle{T2i-adapter: Learning adapters to dig out more controllable ability for text-to-image diffusion models}. In \bibinfo{booktitle}{\emph{Proceedings of the AAAI conference on artificial intelligence}}, Vol.~\bibinfo{volume}{38}. \bibinfo{pages}{4296--4304}.
\newblock


\bibitem[Mueggler et~al\mbox{.}(2017)]%
        {mueggler2017event}
\bibfield{author}{\bibinfo{person}{Elias Mueggler}, \bibinfo{person}{Henri Rebecq}, \bibinfo{person}{Guillermo Gallego}, \bibinfo{person}{Tobi Delbruck}, {and} \bibinfo{person}{Davide Scaramuzza}.} \bibinfo{year}{2017}\natexlab{}.
\newblock \showarticletitle{The event-camera dataset and simulator: Event-based data for pose estimation, visual odometry, and SLAM}.
\newblock \bibinfo{journal}{\emph{The International journal of robotics research}} \bibinfo{volume}{36}, \bibinfo{number}{2} (\bibinfo{year}{2017}), \bibinfo{pages}{142--149}.
\newblock


\bibitem[Munda et~al\mbox{.}(2018)]%
        {munda2018real}
\bibfield{author}{\bibinfo{person}{Gottfried Munda}, \bibinfo{person}{Christian Reinbacher}, {and} \bibinfo{person}{Thomas Pock}.} \bibinfo{year}{2018}\natexlab{}.
\newblock \showarticletitle{Real-time intensity-image reconstruction for event cameras using manifold regularisation}.
\newblock \bibinfo{journal}{\emph{International Journal of Computer Vision}} \bibinfo{volume}{126}, \bibinfo{number}{12} (\bibinfo{year}{2018}), \bibinfo{pages}{1381--1393}.
\newblock


\bibitem[Namekata et~al\mbox{.}(2024)]%
        {namekata2024sg}
\bibfield{author}{\bibinfo{person}{Koichi Namekata}, \bibinfo{person}{Sherwin Bahmani}, \bibinfo{person}{Ziyi Wu}, \bibinfo{person}{Yash Kant}, \bibinfo{person}{Igor Gilitschenski}, {and} \bibinfo{person}{David~B Lindell}.} \bibinfo{year}{2024}\natexlab{}.
\newblock \showarticletitle{Sg-i2v: Self-guided trajectory control in image-to-video generation}.
\newblock \bibinfo{journal}{\emph{arXiv preprint arXiv:2411.04989}} (\bibinfo{year}{2024}).
\newblock


\bibitem[Niu et~al\mbox{.}(2024)]%
        {niu2024mofa}
\bibfield{author}{\bibinfo{person}{Muyao Niu}, \bibinfo{person}{Xiaodong Cun}, \bibinfo{person}{Xintao Wang}, \bibinfo{person}{Yong Zhang}, \bibinfo{person}{Ying Shan}, {and} \bibinfo{person}{Yinqiang Zheng}.} \bibinfo{year}{2024}\natexlab{}.
\newblock \showarticletitle{Mofa-video: Controllable image animation via generative motion field adaptions in frozen image-to-video diffusion model}. In \bibinfo{booktitle}{\emph{European Conference on Computer Vision}}. Springer, \bibinfo{pages}{111--128}.
\newblock


\bibitem[Paikin et~al\mbox{.}(2021)]%
        {paikin2021efi}
\bibfield{author}{\bibinfo{person}{Genady Paikin}, \bibinfo{person}{Yotam Ater}, \bibinfo{person}{Roy Shaul}, {and} \bibinfo{person}{Evgeny Soloveichik}.} \bibinfo{year}{2021}\natexlab{}.
\newblock \showarticletitle{Efi-net: Video frame interpolation from fusion of events and frames}. In \bibinfo{booktitle}{\emph{Proceedings of the IEEE/CVF conference on computer vision and pattern recognition}}. \bibinfo{pages}{1291--1301}.
\newblock


\bibitem[Paredes-Vall{\'e}s and De~Croon(2021)]%
        {paredes2021back}
\bibfield{author}{\bibinfo{person}{Federico Paredes-Vall{\'e}s} {and} \bibinfo{person}{Guido~CHE De~Croon}.} \bibinfo{year}{2021}\natexlab{}.
\newblock \showarticletitle{Back to event basics: Self-supervised learning of image reconstruction for event cameras via photometric constancy}. In \bibinfo{booktitle}{\emph{Proceedings of the IEEE/CVF Conference on Computer Vision and Pattern Recognition}}. \bibinfo{pages}{3446--3455}.
\newblock


\bibitem[Qiu et~al\mbox{.}(2023)]%
        {qiu2023freenoise}
\bibfield{author}{\bibinfo{person}{Haonan Qiu}, \bibinfo{person}{Menghan Xia}, \bibinfo{person}{Yong Zhang}, \bibinfo{person}{Yingqing He}, \bibinfo{person}{Xintao Wang}, \bibinfo{person}{Ying Shan}, {and} \bibinfo{person}{Ziwei Liu}.} \bibinfo{year}{2023}\natexlab{}.
\newblock \showarticletitle{Freenoise: Tuning-free longer video diffusion via noise rescheduling}.
\newblock \bibinfo{journal}{\emph{arXiv preprint arXiv:2310.15169}} (\bibinfo{year}{2023}).
\newblock


\bibitem[Rebecq et~al\mbox{.}(2018)]%
        {rebecq2018esim}
\bibfield{author}{\bibinfo{person}{Henri Rebecq}, \bibinfo{person}{Daniel Gehrig}, {and} \bibinfo{person}{Davide Scaramuzza}.} \bibinfo{year}{2018}\natexlab{}.
\newblock \showarticletitle{Esim: an open event camera simulator}. In \bibinfo{booktitle}{\emph{Conference on robot learning}}. PMLR, \bibinfo{pages}{969--982}.
\newblock


\bibitem[Rebecq et~al\mbox{.}(2019a)]%
        {rebecq2019events}
\bibfield{author}{\bibinfo{person}{Henri Rebecq}, \bibinfo{person}{Ren{\'e} Ranftl}, \bibinfo{person}{Vladlen Koltun}, {and} \bibinfo{person}{Davide Scaramuzza}.} \bibinfo{year}{2019}\natexlab{a}.
\newblock \showarticletitle{Events-to-video: Bringing modern computer vision to event cameras}. In \bibinfo{booktitle}{\emph{Proceedings of the IEEE/CVF Conference on Computer Vision and Pattern Recognition}}. \bibinfo{pages}{3857--3866}.
\newblock


\bibitem[Rebecq et~al\mbox{.}(2019b)]%
        {rebecq2019high}
\bibfield{author}{\bibinfo{person}{Henri Rebecq}, \bibinfo{person}{Ren{\'e} Ranftl}, \bibinfo{person}{Vladlen Koltun}, {and} \bibinfo{person}{Davide Scaramuzza}.} \bibinfo{year}{2019}\natexlab{b}.
\newblock \showarticletitle{High speed and high dynamic range video with an event camera}.
\newblock \bibinfo{journal}{\emph{IEEE transactions on pattern analysis and machine intelligence}} \bibinfo{volume}{43}, \bibinfo{number}{6} (\bibinfo{year}{2019}), \bibinfo{pages}{1964--1980}.
\newblock


\bibitem[Scheerlinck et~al\mbox{.}(2018)]%
        {scheerlinck2018continuous}
\bibfield{author}{\bibinfo{person}{Cedric Scheerlinck}, \bibinfo{person}{Nick Barnes}, {and} \bibinfo{person}{Robert Mahony}.} \bibinfo{year}{2018}\natexlab{}.
\newblock \showarticletitle{Continuous-time intensity estimation using event cameras}. In \bibinfo{booktitle}{\emph{Asian Conference on Computer Vision}}. Springer, \bibinfo{pages}{308--324}.
\newblock


\bibitem[Scheerlinck et~al\mbox{.}(2020)]%
        {scheerlinck2020fast}
\bibfield{author}{\bibinfo{person}{Cedric Scheerlinck}, \bibinfo{person}{Henri Rebecq}, \bibinfo{person}{Daniel Gehrig}, \bibinfo{person}{Nick Barnes}, \bibinfo{person}{Robert Mahony}, {and} \bibinfo{person}{Davide Scaramuzza}.} \bibinfo{year}{2020}\natexlab{}.
\newblock \showarticletitle{Fast image reconstruction with an event camera}. In \bibinfo{booktitle}{\emph{Proceedings of the IEEE/CVF Winter Conference on Applications of Computer Vision}}. \bibinfo{pages}{156--163}.
\newblock


\bibitem[Shi et~al\mbox{.}(2023)]%
        {shi2023ido}
\bibfield{author}{\bibinfo{person}{Chenyang Shi}, \bibinfo{person}{Hanxiao Liu}, \bibinfo{person}{Jing Jin}, \bibinfo{person}{Wenzhuo Li}, \bibinfo{person}{Yuzhen Li}, \bibinfo{person}{Boyi Wei}, {and} \bibinfo{person}{Yibo Zhang}.} \bibinfo{year}{2023}\natexlab{}.
\newblock \showarticletitle{IDO-VFI: Identifying Dynamics via Optical Flow Guidance for Video Frame Interpolation with Events}.
\newblock \bibinfo{journal}{\emph{arXiv preprint arXiv:2305.10198}} (\bibinfo{year}{2023}).
\newblock


\bibitem[Shiu et~al\mbox{.}(2025)]%
        {shiu2025stream}
\bibfield{author}{\bibinfo{person}{Hau-Shiang Shiu}, \bibinfo{person}{Chin-Yang Lin}, \bibinfo{person}{Zhixiang Wang}, \bibinfo{person}{Chi-Wei Hsiao}, \bibinfo{person}{Po-Fan Yu}, \bibinfo{person}{Yu-Chih Chen}, {and} \bibinfo{person}{Yu-Lun Liu}.} \bibinfo{year}{2025}\natexlab{}.
\newblock \showarticletitle{Stream-DiffVSR: Low-Latency Streamable Video Super-Resolution via Auto-Regressive Diffusion}.
\newblock \bibinfo{journal}{\emph{arXiv preprint arXiv:2512.23709}} (\bibinfo{year}{2025}).
\newblock


\bibitem[Singer et~al\mbox{.}(2022)]%
        {singer2022make}
\bibfield{author}{\bibinfo{person}{Uriel Singer}, \bibinfo{person}{Adam Polyak}, \bibinfo{person}{Thomas Hayes}, \bibinfo{person}{Xi Yin}, \bibinfo{person}{Jie An}, \bibinfo{person}{Songyang Zhang}, \bibinfo{person}{Qiyuan Hu}, \bibinfo{person}{Harry Yang}, \bibinfo{person}{Oron Ashual}, \bibinfo{person}{Oran Gafni}, {et~al\mbox{.}}} \bibinfo{year}{2022}\natexlab{}.
\newblock \showarticletitle{Make-a-video: Text-to-video generation without text-video data}.
\newblock \bibinfo{journal}{\emph{arXiv preprint arXiv:2209.14792}} (\bibinfo{year}{2022}).
\newblock


\bibitem[Song et~al\mbox{.}(2020)]%
        {song2020score}
\bibfield{author}{\bibinfo{person}{Yang Song}, \bibinfo{person}{Jascha Sohl-Dickstein}, \bibinfo{person}{Diederik~P Kingma}, \bibinfo{person}{Abhishek Kumar}, \bibinfo{person}{Stefano Ermon}, {and} \bibinfo{person}{Ben Poole}.} \bibinfo{year}{2020}\natexlab{}.
\newblock \showarticletitle{Score-based generative modeling through stochastic differential equations}.
\newblock \bibinfo{journal}{\emph{arXiv preprint arXiv:2011.13456}} (\bibinfo{year}{2020}).
\newblock


\bibitem[Stoffregen et~al\mbox{.}(2020)]%
        {stoffregen2020reducing}
\bibfield{author}{\bibinfo{person}{Timo Stoffregen}, \bibinfo{person}{Cedric Scheerlinck}, \bibinfo{person}{Davide Scaramuzza}, \bibinfo{person}{Tom Drummond}, \bibinfo{person}{Nick Barnes}, \bibinfo{person}{Lindsay Kleeman}, {and} \bibinfo{person}{Robert Mahony}.} \bibinfo{year}{2020}\natexlab{}.
\newblock \showarticletitle{Reducing the sim-to-real gap for event cameras}. In \bibinfo{booktitle}{\emph{European Conference on Computer Vision}}. Springer, \bibinfo{pages}{534--549}.
\newblock


\bibitem[Sun et~al\mbox{.}(2024)]%
        {sun2024unified}
\bibfield{author}{\bibinfo{person}{Lei Sun}, \bibinfo{person}{Daniel Gehrig}, \bibinfo{person}{Christos Sakaridis}, \bibinfo{person}{Mathias Gehrig}, \bibinfo{person}{Jingyun Liang}, \bibinfo{person}{Peng Sun}, \bibinfo{person}{Zhijie Xu}, \bibinfo{person}{Kaiwei Wang}, \bibinfo{person}{Luc Van~Gool}, {and} \bibinfo{person}{Davide Scaramuzza}.} \bibinfo{year}{2024}\natexlab{}.
\newblock \showarticletitle{A unified framework for event-based frame interpolation with ad-hoc deblurring in the wild}.
\newblock \bibinfo{journal}{\emph{IEEE Transactions on Pattern Analysis and Machine Intelligence}} (\bibinfo{year}{2024}).
\newblock


\bibitem[Sun et~al\mbox{.}(2023)]%
        {sun2023event}
\bibfield{author}{\bibinfo{person}{Lei Sun}, \bibinfo{person}{Christos Sakaridis}, \bibinfo{person}{Jingyun Liang}, \bibinfo{person}{Peng Sun}, \bibinfo{person}{Jiezhang Cao}, \bibinfo{person}{Kai Zhang}, \bibinfo{person}{Qi Jiang}, \bibinfo{person}{Kaiwei Wang}, {and} \bibinfo{person}{Luc Van~Gool}.} \bibinfo{year}{2023}\natexlab{}.
\newblock \showarticletitle{Event-based frame interpolation with ad-hoc deblurring}. In \bibinfo{booktitle}{\emph{Proceedings of the IEEE/CVF Conference on Computer Vision and Pattern Recognition}}. \bibinfo{pages}{18043--18052}.
\newblock


\bibitem[Team(2024)]%
        {genmo2024mochi}
\bibfield{author}{\bibinfo{person}{Genmo Team}.} \bibinfo{year}{2024}\natexlab{}.
\newblock \bibinfo{title}{Mochi 1}.
\newblock \bibinfo{howpublished}{\url{https://github.com/genmoai/models}}.
\newblock


\bibitem[Tulyakov et~al\mbox{.}(2022)]%
        {tulyakov2022time}
\bibfield{author}{\bibinfo{person}{Stepan Tulyakov}, \bibinfo{person}{Alfredo Bochicchio}, \bibinfo{person}{Daniel Gehrig}, \bibinfo{person}{Stamatios Georgoulis}, \bibinfo{person}{Yuanyou Li}, {and} \bibinfo{person}{Davide Scaramuzza}.} \bibinfo{year}{2022}\natexlab{}.
\newblock \showarticletitle{Time lens++: Event-based frame interpolation with parametric non-linear flow and multi-scale fusion}. In \bibinfo{booktitle}{\emph{Proceedings of the IEEE/CVF Conference on Computer Vision and Pattern Recognition}}. \bibinfo{pages}{17755--17764}.
\newblock


\bibitem[Tulyakov et~al\mbox{.}(2021)]%
        {tulyakov2021time}
\bibfield{author}{\bibinfo{person}{Stepan Tulyakov}, \bibinfo{person}{Daniel Gehrig}, \bibinfo{person}{Stamatios Georgoulis}, \bibinfo{person}{Julius Erbach}, \bibinfo{person}{Mathias Gehrig}, \bibinfo{person}{Yuanyou Li}, {and} \bibinfo{person}{Davide Scaramuzza}.} \bibinfo{year}{2021}\natexlab{}.
\newblock \showarticletitle{Time lens: Event-based video frame interpolation}. In \bibinfo{booktitle}{\emph{Proceedings of the IEEE/CVF conference on computer vision and pattern recognition}}. \bibinfo{pages}{16155--16164}.
\newblock


\bibitem[Wan et~al\mbox{.}(2025)]%
        {wan2025wan}
\bibfield{author}{\bibinfo{person}{Team Wan}, \bibinfo{person}{Ang Wang}, \bibinfo{person}{Baole Ai}, \bibinfo{person}{Bin Wen}, \bibinfo{person}{Chaojie Mao}, \bibinfo{person}{Chen-Wei Xie}, \bibinfo{person}{Di Chen}, \bibinfo{person}{Feiwu Yu}, \bibinfo{person}{Haiming Zhao}, \bibinfo{person}{Jianxiao Yang}, {et~al\mbox{.}}} \bibinfo{year}{2025}\natexlab{}.
\newblock \showarticletitle{Wan: Open and advanced large-scale video generative models}.
\newblock \bibinfo{journal}{\emph{arXiv preprint arXiv:2503.20314}} (\bibinfo{year}{2025}).
\newblock


\bibitem[Wang et~al\mbox{.}(2024a)]%
        {wang2024easycontrol}
\bibfield{author}{\bibinfo{person}{Cong Wang}, \bibinfo{person}{Jiaxi Gu}, \bibinfo{person}{Panwen Hu}, \bibinfo{person}{Haoyu Zhao}, \bibinfo{person}{Yuanfan Guo}, \bibinfo{person}{Jianhua Han}, \bibinfo{person}{Hang Xu}, {and} \bibinfo{person}{Xiaodan Liang}.} \bibinfo{year}{2024}\natexlab{a}.
\newblock \showarticletitle{Easycontrol: Transfer controlnet to video diffusion for controllable generation and interpolation}.
\newblock \bibinfo{journal}{\emph{arXiv preprint arXiv:2408.13005}} (\bibinfo{year}{2024}).
\newblock


\bibitem[Wang et~al\mbox{.}(2024d)]%
        {wang2024boximator}
\bibfield{author}{\bibinfo{person}{Jiawei Wang}, \bibinfo{person}{Yuchen Zhang}, \bibinfo{person}{Jiaxin Zou}, \bibinfo{person}{Yan Zeng}, \bibinfo{person}{Guoqiang Wei}, \bibinfo{person}{Liping Yuan}, {and} \bibinfo{person}{Hang Li}.} \bibinfo{year}{2024}\natexlab{d}.
\newblock \showarticletitle{Boximator: Generating rich and controllable motions for video synthesis}.
\newblock \bibinfo{journal}{\emph{arXiv preprint arXiv:2402.01566}} (\bibinfo{year}{2024}).
\newblock


\bibitem[Wang et~al\mbox{.}(2019)]%
        {wang2019event}
\bibfield{author}{\bibinfo{person}{Lin Wang}, \bibinfo{person}{Yo-Sung Ho}, \bibinfo{person}{Kuk-Jin Yoon}, {et~al\mbox{.}}} \bibinfo{year}{2019}\natexlab{}.
\newblock \showarticletitle{Event-based high dynamic range image and very high frame rate video generation using conditional generative adversarial networks}. In \bibinfo{booktitle}{\emph{Proceedings of the IEEE/CVF Conference on Computer Vision and Pattern Recognition}}. \bibinfo{pages}{10081--10090}.
\newblock


\bibitem[Wang et~al\mbox{.}(2023)]%
        {wang2023videocomposer}
\bibfield{author}{\bibinfo{person}{Xiang Wang}, \bibinfo{person}{Hangjie Yuan}, \bibinfo{person}{Shiwei Zhang}, \bibinfo{person}{Dayou Chen}, \bibinfo{person}{Jiuniu Wang}, \bibinfo{person}{Yingya Zhang}, \bibinfo{person}{Yujun Shen}, \bibinfo{person}{Deli Zhao}, {and} \bibinfo{person}{Jingren Zhou}.} \bibinfo{year}{2023}\natexlab{}.
\newblock \showarticletitle{Videocomposer: Compositional video synthesis with motion controllability}.
\newblock \bibinfo{journal}{\emph{Advances in Neural Information Processing Systems}}  \bibinfo{volume}{36} (\bibinfo{year}{2023}), \bibinfo{pages}{7594--7611}.
\newblock


\bibitem[Wang et~al\mbox{.}(2024b)]%
        {wang2024revisit}
\bibfield{author}{\bibinfo{person}{Zipeng Wang}, \bibinfo{person}{Yunfan Lu}, {and} \bibinfo{person}{Lin Wang}.} \bibinfo{year}{2024}\natexlab{b}.
\newblock \showarticletitle{Revisit event generation model: Self-supervised learning of event-to-video reconstruction with implicit neural representations}. In \bibinfo{booktitle}{\emph{European Conference on Computer Vision}}. Springer, \bibinfo{pages}{321--339}.
\newblock


\bibitem[Wang et~al\mbox{.}(2024c)]%
        {wang2024motionctrl}
\bibfield{author}{\bibinfo{person}{Zhouxia Wang}, \bibinfo{person}{Ziyang Yuan}, \bibinfo{person}{Xintao Wang}, \bibinfo{person}{Yaowei Li}, \bibinfo{person}{Tianshui Chen}, \bibinfo{person}{Menghan Xia}, \bibinfo{person}{Ping Luo}, {and} \bibinfo{person}{Ying Shan}.} \bibinfo{year}{2024}\natexlab{c}.
\newblock \showarticletitle{Motionctrl: A unified and flexible motion controller for video generation}. In \bibinfo{booktitle}{\emph{ACM SIGGRAPH 2024 Conference Papers}}. \bibinfo{pages}{1--11}.
\newblock


\bibitem[Weng et~al\mbox{.}(2021)]%
        {weng2021event}
\bibfield{author}{\bibinfo{person}{Wenming Weng}, \bibinfo{person}{Yueyi Zhang}, {and} \bibinfo{person}{Zhiwei Xiong}.} \bibinfo{year}{2021}\natexlab{}.
\newblock \showarticletitle{Event-based video reconstruction using transformer}. In \bibinfo{booktitle}{\emph{Proceedings of the IEEE/CVF International Conference on Computer Vision}}. \bibinfo{pages}{2563--2572}.
\newblock


\bibitem[Wu et~al\mbox{.}(2022)]%
        {wu2022video}
\bibfield{author}{\bibinfo{person}{Song Wu}, \bibinfo{person}{Kaichao You}, \bibinfo{person}{Weihua He}, \bibinfo{person}{Chen Yang}, \bibinfo{person}{Yang Tian}, \bibinfo{person}{Yaoyuan Wang}, \bibinfo{person}{Ziyang Zhang}, {and} \bibinfo{person}{Jianxing Liao}.} \bibinfo{year}{2022}\natexlab{}.
\newblock \showarticletitle{Video interpolation by event-driven anisotropic adjustment of optical flow}. In \bibinfo{booktitle}{\emph{European Conference on Computer Vision}}. Springer, \bibinfo{pages}{267--283}.
\newblock


\bibitem[Wu et~al\mbox{.}(2024)]%
        {wu2024draganything}
\bibfield{author}{\bibinfo{person}{Weijia Wu}, \bibinfo{person}{Zhuang Li}, \bibinfo{person}{Yuchao Gu}, \bibinfo{person}{Rui Zhao}, \bibinfo{person}{Yefei He}, \bibinfo{person}{David~Junhao Zhang}, \bibinfo{person}{Mike~Zheng Shou}, \bibinfo{person}{Yan Li}, \bibinfo{person}{Tingting Gao}, {and} \bibinfo{person}{Di Zhang}.} \bibinfo{year}{2024}\natexlab{}.
\newblock \showarticletitle{Draganything: Motion control for anything using entity representation}. In \bibinfo{booktitle}{\emph{European Conference on Computer Vision}}. Springer, \bibinfo{pages}{331--348}.
\newblock


\bibitem[Xiao et~al\mbox{.}(2024)]%
        {xiao2024video}
\bibfield{author}{\bibinfo{person}{Zeqi Xiao}, \bibinfo{person}{Yifan Zhou}, \bibinfo{person}{Shuai Yang}, {and} \bibinfo{person}{Xingang Pan}.} \bibinfo{year}{2024}\natexlab{}.
\newblock \showarticletitle{Video diffusion models are training-free motion interpreter and controller}.
\newblock \bibinfo{journal}{\emph{Advances in Neural Information Processing Systems}}  \bibinfo{volume}{37} (\bibinfo{year}{2024}), \bibinfo{pages}{76115--76138}.
\newblock


\bibitem[Yang et~al\mbox{.}(2024)]%
        {yang2024cogvideox}
\bibfield{author}{\bibinfo{person}{Zhuoyi Yang}, \bibinfo{person}{Jiayan Teng}, \bibinfo{person}{Wendi Zheng}, \bibinfo{person}{Ming Ding}, \bibinfo{person}{Shiyu Huang}, \bibinfo{person}{Jiazheng Xu}, \bibinfo{person}{Yuanming Yang}, \bibinfo{person}{Wenyi Hong}, \bibinfo{person}{Xiaohan Zhang}, \bibinfo{person}{Guanyu Feng}, {et~al\mbox{.}}} \bibinfo{year}{2024}\natexlab{}.
\newblock \showarticletitle{Cogvideox: Text-to-video diffusion models with an expert transformer}.
\newblock \bibinfo{journal}{\emph{arXiv preprint arXiv:2408.06072}} (\bibinfo{year}{2024}).
\newblock


\bibitem[Yeh et~al\mbox{.}(2024)]%
        {yeh2024diffir2vr}
\bibfield{author}{\bibinfo{person}{Chang-Han Yeh}, \bibinfo{person}{Hau-Shiang Shiu}, \bibinfo{person}{Chin-Yang Lin}, \bibinfo{person}{Zhixiang Wang}, \bibinfo{person}{Chi-Wei Hsiao}, \bibinfo{person}{Ting-Hsuan Chen}, {and} \bibinfo{person}{Yu-Lun Liu}.} \bibinfo{year}{2024}\natexlab{}.
\newblock \showarticletitle{Diffir2vr-zero: Zero-shot video restoration with diffusion-based image restoration models}.
\newblock \bibinfo{journal}{\emph{arXiv preprint arXiv:2407.01519}} (\bibinfo{year}{2024}).
\newblock


\bibitem[Yin et~al\mbox{.}(2023)]%
        {yin2023nuwa}
\bibfield{author}{\bibinfo{person}{Shengming Yin}, \bibinfo{person}{Chenfei Wu}, \bibinfo{person}{Huan Yang}, \bibinfo{person}{Jianfeng Wang}, \bibinfo{person}{Xiaodong Wang}, \bibinfo{person}{Minheng Ni}, \bibinfo{person}{Zhengyuan Yang}, \bibinfo{person}{Linjie Li}, \bibinfo{person}{Shuguang Liu}, \bibinfo{person}{Fan Yang}, {et~al\mbox{.}}} \bibinfo{year}{2023}\natexlab{}.
\newblock \showarticletitle{Nuwa-xl: Diffusion over diffusion for extremely long video generation}. In \bibinfo{booktitle}{\emph{Proceedings of the 61st Annual Meeting of the Association for Computational Linguistics (Volume 1: Long Papers)}}. \bibinfo{pages}{1309--1320}.
\newblock


\bibitem[Yin et~al\mbox{.}(2025)]%
        {yin2025slow}
\bibfield{author}{\bibinfo{person}{Tianwei Yin}, \bibinfo{person}{Qiang Zhang}, \bibinfo{person}{Richard Zhang}, \bibinfo{person}{William~T Freeman}, \bibinfo{person}{Fredo Durand}, \bibinfo{person}{Eli Shechtman}, {and} \bibinfo{person}{Xun Huang}.} \bibinfo{year}{2025}\natexlab{}.
\newblock \showarticletitle{From slow bidirectional to fast autoregressive video diffusion models}. In \bibinfo{booktitle}{\emph{Proceedings of the Computer Vision and Pattern Recognition Conference}}. \bibinfo{pages}{22963--22974}.
\newblock


\bibitem[Yoo et~al\mbox{.}(2023)]%
        {yoo2023towards}
\bibfield{author}{\bibinfo{person}{Jaehoon Yoo}, \bibinfo{person}{Semin Kim}, \bibinfo{person}{Doyup Lee}, \bibinfo{person}{Chiheon Kim}, {and} \bibinfo{person}{Seunghoon Hong}.} \bibinfo{year}{2023}\natexlab{}.
\newblock \showarticletitle{Towards end-to-end generative modeling of long videos with memory-efficient bidirectional transformers}. In \bibinfo{booktitle}{\emph{Proceedings of the IEEE/CVF Conference on Computer Vision and Pattern Recognition}}. \bibinfo{pages}{22888--22897}.
\newblock


\bibitem[Zhang et~al\mbox{.}(2025b)]%
        {zhang2025evdi++}
\bibfield{author}{\bibinfo{person}{Chi Zhang}, \bibinfo{person}{Xiang Zhang}, \bibinfo{person}{Chenxu Jiang}, \bibinfo{person}{Gui-Song Xia}, {and} \bibinfo{person}{Lei Yu}.} \bibinfo{year}{2025}\natexlab{b}.
\newblock \showarticletitle{EVDI++: Event-based Video Deblurring and Interpolation via Self-Supervised Learning}.
\newblock \bibinfo{journal}{\emph{arXiv preprint arXiv:2509.08260}} (\bibinfo{year}{2025}).
\newblock


\bibitem[Zhang and Agrawala(2025)]%
        {zhang2025packing}
\bibfield{author}{\bibinfo{person}{Lvmin Zhang} {and} \bibinfo{person}{Maneesh Agrawala}.} \bibinfo{year}{2025}\natexlab{}.
\newblock \showarticletitle{Packing input frame context in next-frame prediction models for video generation}.
\newblock \bibinfo{journal}{\emph{arXiv preprint arXiv:2504.12626}} (\bibinfo{year}{2025}).
\newblock


\bibitem[Zhang et~al\mbox{.}(2023)]%
        {zhang2023adding}
\bibfield{author}{\bibinfo{person}{Lvmin Zhang}, \bibinfo{person}{Anyi Rao}, {and} \bibinfo{person}{Maneesh Agrawala}.} \bibinfo{year}{2023}\natexlab{}.
\newblock \showarticletitle{Adding conditional control to text-to-image diffusion models}. In \bibinfo{booktitle}{\emph{Proceedings of the IEEE/CVF international conference on computer vision}}. \bibinfo{pages}{3836--3847}.
\newblock


\bibitem[Zhang and Yu(2022)]%
        {zhang2022unifying}
\bibfield{author}{\bibinfo{person}{Xiang Zhang} {and} \bibinfo{person}{Lei Yu}.} \bibinfo{year}{2022}\natexlab{}.
\newblock \showarticletitle{Unifying motion deblurring and frame interpolation with events}. In \bibinfo{booktitle}{\emph{Proceedings of the IEEE/CVF Conference on Computer Vision and Pattern Recognition}}. \bibinfo{pages}{17765--17774}.
\newblock


\bibitem[Zhang et~al\mbox{.}(2025a)]%
        {zhang2025tora}
\bibfield{author}{\bibinfo{person}{Zhenghao Zhang}, \bibinfo{person}{Junchao Liao}, \bibinfo{person}{Menghao Li}, \bibinfo{person}{Zuozhuo Dai}, \bibinfo{person}{Bingxue Qiu}, \bibinfo{person}{Siyu Zhu}, \bibinfo{person}{Long Qin}, {and} \bibinfo{person}{Weizhi Wang}.} \bibinfo{year}{2025}\natexlab{a}.
\newblock \showarticletitle{Tora: Trajectory-oriented diffusion transformer for video generation}. In \bibinfo{booktitle}{\emph{Proceedings of the Computer Vision and Pattern Recognition Conference}}. \bibinfo{pages}{2063--2073}.
\newblock


\bibitem[Zhang et~al\mbox{.}(2022)]%
        {zhang2022formulating}
\bibfield{author}{\bibinfo{person}{Zelin Zhang}, \bibinfo{person}{Anthony~J Yezzi}, {and} \bibinfo{person}{Guillermo Gallego}.} \bibinfo{year}{2022}\natexlab{}.
\newblock \showarticletitle{Formulating event-based image reconstruction as a linear inverse problem with deep regularization using optical flow}.
\newblock \bibinfo{journal}{\emph{IEEE Transactions on Pattern Analysis and Machine Intelligence}} \bibinfo{volume}{45}, \bibinfo{number}{7} (\bibinfo{year}{2022}), \bibinfo{pages}{8372--8389}.
\newblock


\bibitem[Zhao et~al\mbox{.}(2024)]%
        {zhao2024moviedreamer}
\bibfield{author}{\bibinfo{person}{Canyu Zhao}, \bibinfo{person}{Mingyu Liu}, \bibinfo{person}{Wen Wang}, \bibinfo{person}{Weihua Chen}, \bibinfo{person}{Fan Wang}, \bibinfo{person}{Hao Chen}, \bibinfo{person}{Bo Zhang}, {and} \bibinfo{person}{Chunhua Shen}.} \bibinfo{year}{2024}\natexlab{}.
\newblock \showarticletitle{Moviedreamer: Hierarchical generation for coherent long visual sequence}.
\newblock \bibinfo{journal}{\emph{arXiv preprint arXiv:2407.16655}} (\bibinfo{year}{2024}).
\newblock


\bibitem[Zheng et~al\mbox{.}(2024)]%
        {zheng2024open}
\bibfield{author}{\bibinfo{person}{Zangwei Zheng}, \bibinfo{person}{Xiangyu Peng}, \bibinfo{person}{Tianji Yang}, \bibinfo{person}{Chenhui Shen}, \bibinfo{person}{Shenggui Li}, \bibinfo{person}{Hongxin Liu}, \bibinfo{person}{Yukun Zhou}, \bibinfo{person}{Tianyi Li}, {and} \bibinfo{person}{Yang You}.} \bibinfo{year}{2024}\natexlab{}.
\newblock \showarticletitle{Open-sora: Democratizing efficient video production for all}.
\newblock \bibinfo{journal}{\emph{arXiv preprint arXiv:2412.20404}} (\bibinfo{year}{2024}).
\newblock


\bibitem[Zhou et~al\mbox{.}(2024)]%
        {zhou2024upscale}
\bibfield{author}{\bibinfo{person}{Shangchen Zhou}, \bibinfo{person}{Peiqing Yang}, \bibinfo{person}{Jianyi Wang}, \bibinfo{person}{Yihang Luo}, {and} \bibinfo{person}{Chen~Change Loy}.} \bibinfo{year}{2024}\natexlab{}.
\newblock \showarticletitle{Upscale-a-video: Temporal-consistent diffusion model for real-world video super-resolution}. In \bibinfo{booktitle}{\emph{Proceedings of the IEEE/CVF Conference on Computer Vision and Pattern Recognition}}. \bibinfo{pages}{2535--2545}.
\newblock


\bibitem[Zhu et~al\mbox{.}(2018)]%
        {zhu2018multivehicle}
\bibfield{author}{\bibinfo{person}{Alex~Zihao Zhu}, \bibinfo{person}{Dinesh Thakur}, \bibinfo{person}{Tolga {\"O}zaslan}, \bibinfo{person}{Bernd Pfrommer}, \bibinfo{person}{Vijay Kumar}, {and} \bibinfo{person}{Kostas Daniilidis}.} \bibinfo{year}{2018}\natexlab{}.
\newblock \showarticletitle{The multivehicle stereo event camera dataset: An event camera dataset for 3D perception}.
\newblock \bibinfo{journal}{\emph{IEEE Robotics and Automation Letters}} \bibinfo{volume}{3}, \bibinfo{number}{3} (\bibinfo{year}{2018}), \bibinfo{pages}{2032--2039}.
\newblock


\bibitem[Zhu et~al\mbox{.}(2020)]%
        {zhu2020retina}
\bibfield{author}{\bibinfo{person}{Lin Zhu}, \bibinfo{person}{Siwei Dong}, \bibinfo{person}{Jianing Li}, \bibinfo{person}{Tiejun Huang}, {and} \bibinfo{person}{Yonghong Tian}.} \bibinfo{year}{2020}\natexlab{}.
\newblock \showarticletitle{Retina-like visual image reconstruction via spiking neural model}. In \bibinfo{booktitle}{\emph{Proceedings of the IEEE/CVF Conference on Computer Vision and Pattern Recognition}}. \bibinfo{pages}{1438--1446}.
\newblock


\bibitem[Zhu et~al\mbox{.}(2022)]%
        {zhu2022event}
\bibfield{author}{\bibinfo{person}{Lin Zhu}, \bibinfo{person}{Xiao Wang}, \bibinfo{person}{Yi Chang}, \bibinfo{person}{Jianing Li}, \bibinfo{person}{Tiejun Huang}, {and} \bibinfo{person}{Yonghong Tian}.} \bibinfo{year}{2022}\natexlab{}.
\newblock \showarticletitle{Event-based video reconstruction via potential-assisted spiking neural network}. In \bibinfo{booktitle}{\emph{Proceedings of the IEEE/CVF conference on computer vision and pattern recognition}}. \bibinfo{pages}{3594--3604}.
\newblock


\bibitem[Zhu et~al\mbox{.}(2024)]%
        {zhu2024temporal}
\bibfield{author}{\bibinfo{person}{Lin Zhu}, \bibinfo{person}{Yunlong Zheng}, \bibinfo{person}{Yijun Zhang}, \bibinfo{person}{Xiao Wang}, \bibinfo{person}{Lizhi Wang}, {and} \bibinfo{person}{Hua Huang}.} \bibinfo{year}{2024}\natexlab{}.
\newblock \showarticletitle{Temporal Residual Guided Diffusion Framework for Event-Driven Video Reconstruction}. In \bibinfo{booktitle}{\emph{European Conference on Computer Vision}}. Springer, \bibinfo{pages}{411--427}.
\newblock


\end{thebibliography}

%%
%% If your work has an appendix, this is the place to put it.
\appendix

\begin{figure*}
  \includegraphics[width=0.93\textwidth]{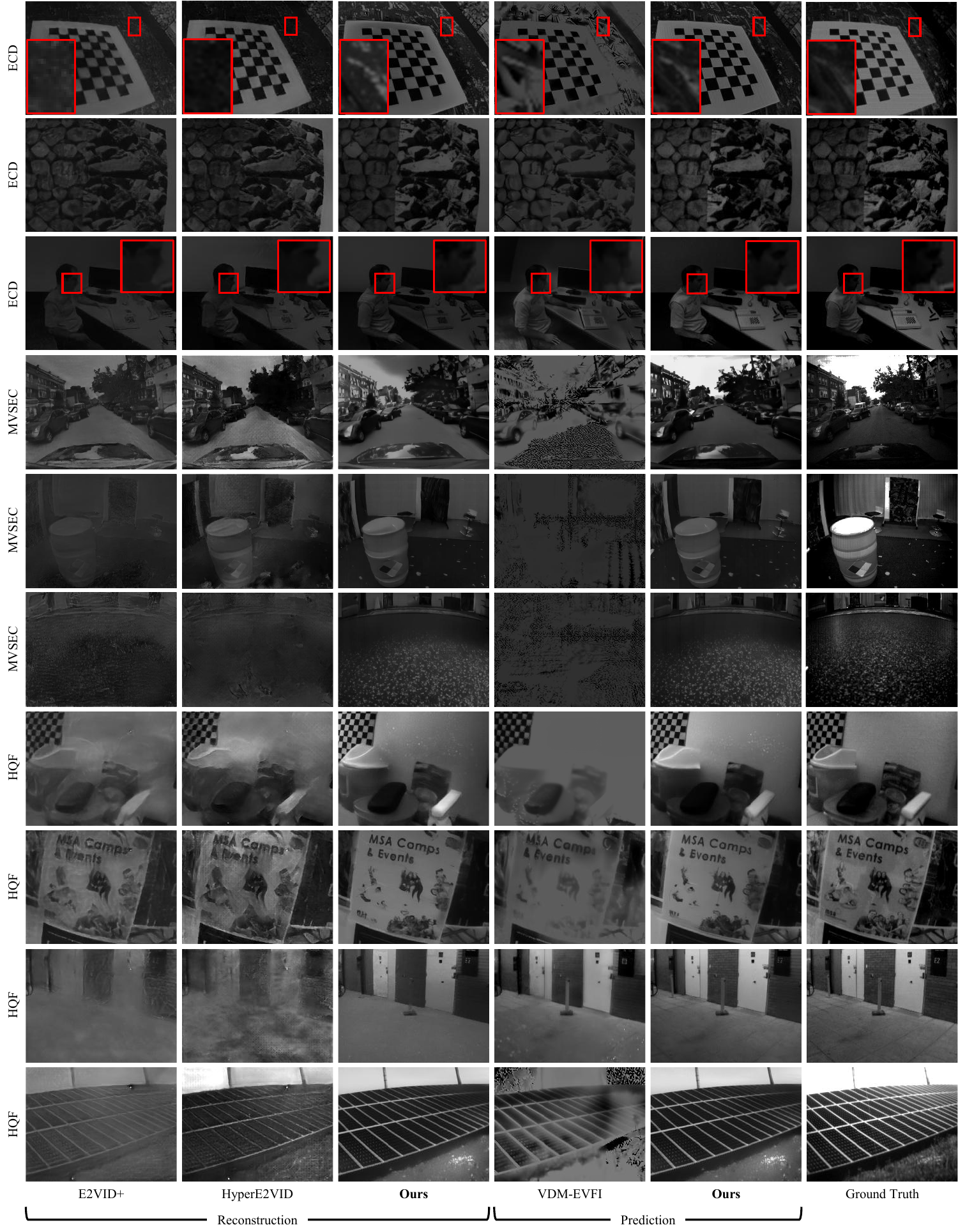}
  \caption{
  \textbf{Additional qualitative comparisons on ECD~\cite{mueggler2017event}, MVSEC~\cite{zhu2018multivehicle}, and HQF datasets~\cite{stoffregen2020reducing}.}
  }
  \label{fig:appendix_recon}
\end{figure*}

\begin{figure*}
  \includegraphics[width=0.98\textwidth]{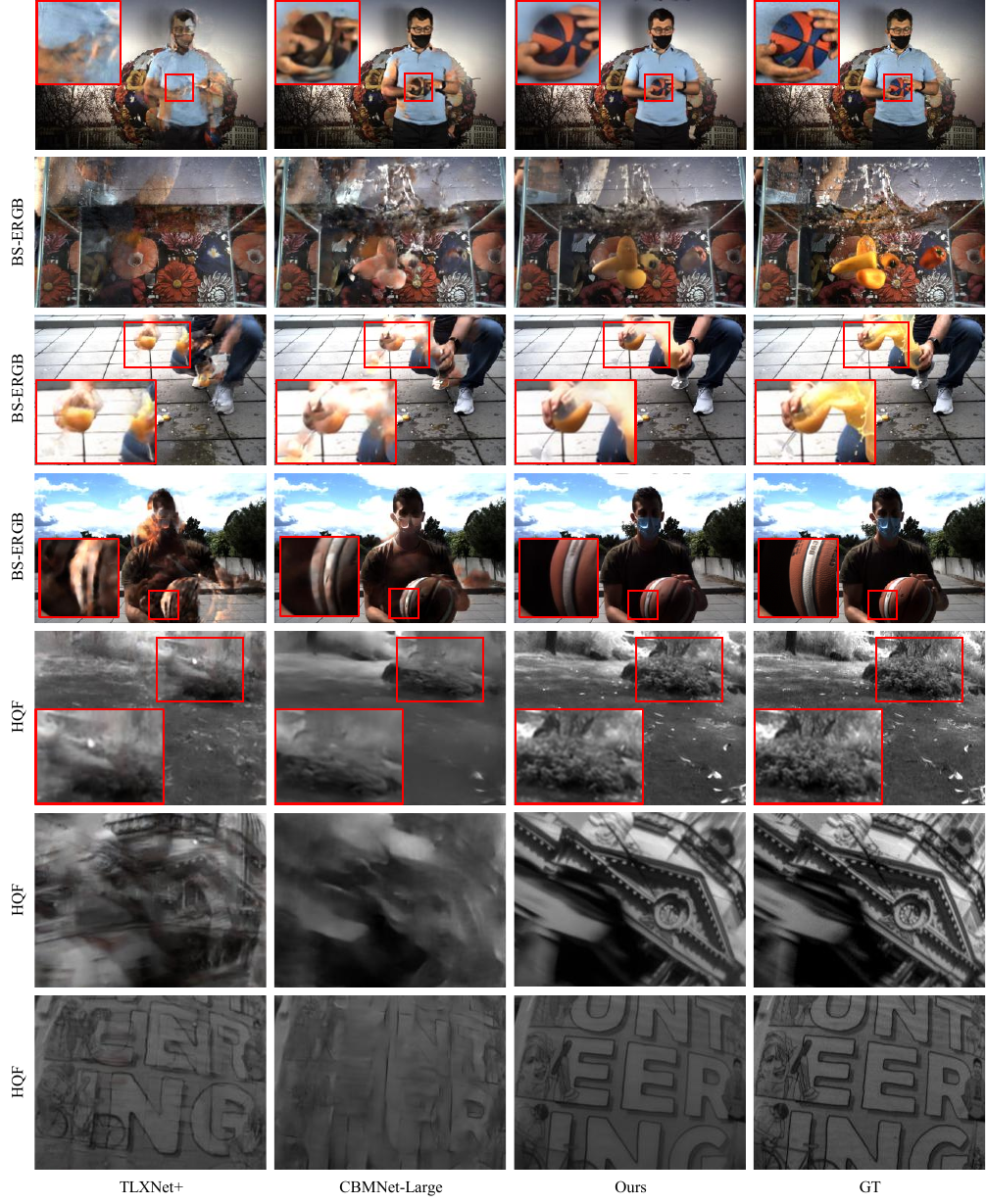}
  \caption{
  \textbf{Additional zero-shot interpolation results on BS-ERGB~\cite{tulyakov2021time} and HQF~\cite{stoffregen2020reducing} datasets.}
  }
  \label{fig:appendix_interpolation}
\end{figure*}

% \section{Supplementary}
\section{Limitations}
\label{sec:limitations}
\begin{figure*}
  \includegraphics[width=\textwidth]{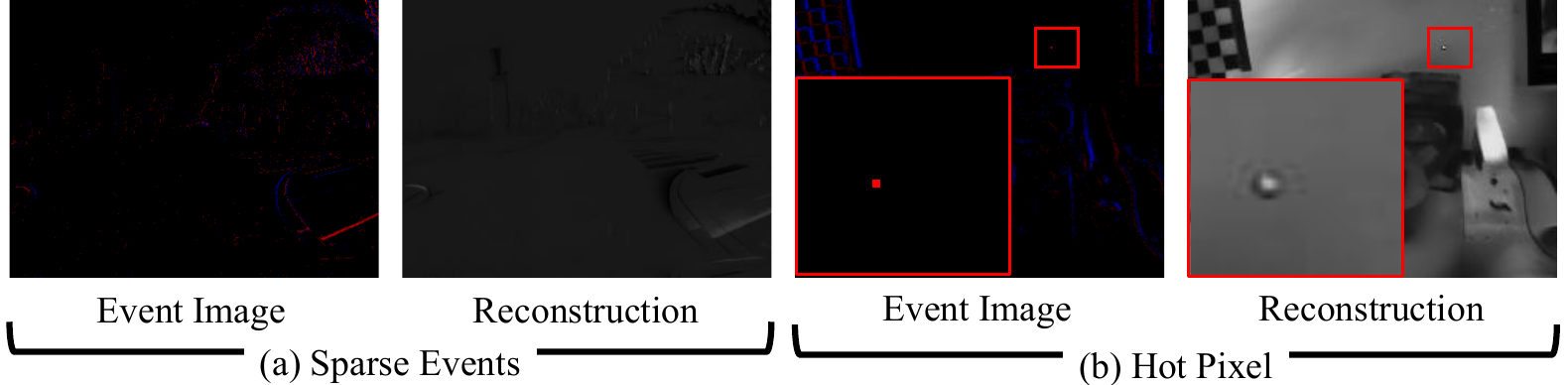}
  \caption{
\textbf{Limitations.} (a) Failure cases under sparse or low-quality event streams. (b) Sensitivity to noise where "hot pixel" is preserved or amplified in the reconstructed frames.
}
\label{fig:limitations}
\end{figure*}

When the input event streams are highly sparse or of poor quality, our method struggles to reconstruct high-quality frames, as illustrated in Fig.~\ref{fig:limitations} (a). Furthermore, Fig.~\ref{fig:limitations} (b) demonstrates that our approach is sensitive to noise in the event condition; specifically, "hot pixel" is often erroneously preserved or amplified in the reconstructed frames.

\section{Ablation of Different Backbones}
To demonstrate the effectiveness of our approach, we integrated it with the Wan 2.2 5B model and conducted an ablation study following the same setup as in Tab.~\ref{tab:ablation_recon}. As shown in Tab.~\ref{tab:ablation_different}, the incremental addition of each component leads to consistent performance gains, proving that our method is effective across different backbones.

\begin{table}[h]
\centering
% \small
\caption{
\textbf{Ablation of Different Backbones on HQF dataset.}
}
\label{tab:ablation_different}
    \resizebox{\columnwidth}{!}{%
\begin{tabular}{cccc|ccc}
    \toprule
    Pretrained & & AR & Adaptive \\
    prior &  Context &  unroll. & ctx. switch  & PSNR↑ & SSIM↑ & LPIPS↓ \\
    \midrule
    & \checkmark  &\checkmark &\checkmark &9.29  &0.032 &0.687  \\
    \checkmark & &  & &13.97  &0.426  &0.395  \\
    \checkmark & \checkmark & & &15.20 &0.534  &0.316 \\
    \checkmark & \checkmark & \checkmark  &  &\textbf{15.55}  &0.526  &0.304  \\
    \checkmark & \checkmark & \checkmark  &\checkmark  &15.50  &\textbf{0.549}  &\textbf{0.286}  \\
    \bottomrule
\end{tabular}
}
\end{table}

\section{Long-term temporal consistency.}
To verify that our improvements stem from long-term stability rather than solely from the DiT backbone's image quality enhancements, we evaluate subject consistency using VBench. As shown in Tab. \ref{tab:vbench_results}, both our reconstruction and prediction variants significantly outperform existing methods. Notably, Ours (Recon) achieves a score of 0.7204, a substantial margin over VDM-EVFI. These results demonstrate that our framework effectively maintains identity and structural integrity across extended sequences, successfully mitigating temporal drifting.
\begin{table}[h]
\centering
\caption{\textbf{Quantitative comparison on VBench Subject Consistency on HQF dataset.}}
\label{tab:vbench_results}
\begin{tabular}{lc}
\toprule
Method & Subject Consistency $\uparrow$ \\
\midrule
E2VID+ (Recon) & 0.5413 \\
HyperE2VID (Recon) & 0.4953 \\
\textbf{Ours (Recon)} & \textbf{0.7204} \\
VDM-EVFI (Pred) & 0.6279 \\
\textbf{Ours (Pred)} & \textbf{0.7187} \\
\bottomrule
\end{tabular}
\end{table}

\section{More Implementation Details.}
We employ CogVideoX I2V as our backbone to generate 49-frame clips at a resolution of 720 × 480, applying Low-Rank Adaptation (LoRA) with a rank of $r = 64$ to the DiT blocks while fully fine-tuning the first projection layer. The model is trained on a NVIDIA RTX PRO 6000 GPU using the AdamW optimizer with a cosine learning rate scheduler, a learning rate of 0.003, a weight decay of 0.01, a batch size of 1, and gradient accumulation over 4 steps. Our training strategy incorporates Autoregressive Unrolling, beginning with 3,000 steps on ground-truth context followed by three iterative cycles where the model is trained for 3,000 steps per cycle using its own previous inferences as context.

During inference, we maintain a 20-frame context. Specifically, for the first chunk during inference where historical context is unavailable, we apply task-specific initialization: reconstruction employs zero tensors for both the start frame latent and context latents; prediction replicates the starting frame $20\times$ to populate the context; and frame interpolation replicates the start and end frames $10\times$ each to form the context. Across all tasks, the corresponding context event voxels are consistently zero-padded. Starting from the second chunk in reconstruction and prediction tasks, the last frame of the previous chunk serves as the first frame for the current one until the sequence is complete. For all reconstruction and prediction experiments of our method, the Adaptive Context Switching threshold is set to 0.05. In the frame interpolation task, we perform alpha blending to fuse forward and backward latents, using a linear weight transition from $1:0$ to $0:1$ that is proportional to the temporal distance from the start and end frames. Finally, for frame interpolation on the BS-ERGB dataset, we follow the Per-tile Denoising and Fusion strategy from VDM-EVFI, upsampling the input to 1952 × 1264—approximately double the original resolution—to preserve fine spatial details in the start and end frames. Although our model inherently produces RGB outputs, we convert these results to grayscale during evaluation to align with the ground-truth format of specific datasets and ensure a fair comparison.

\section{Inference Speed}
To evaluate computational efficiency, we compare the inference speed of our method with E2VID (for reconstruction) and VDM-EVFI (for prediction). For our method and VDM-EVFI, the inference speed is calculated by dividing the time required to generate a single chunk by the number of frames produced in that generation. Specifically, our method generates 49 frames per chunk, whereas VDM-EVFI generates 13 frames per chunk. For E2VID, the speed is measured based on a single frame generation time. The experiments are conducted on an NVIDIA RTX A6000 GPU, excluding pre-processing and post-processing times. In this experiment, we evaluate the performance on the \textit{bike\_bay\_hdr} sequence from the HQF dataset. To ensure a steady-state measurement and account for initialization overhead, we report the generation speed of the second chunk for our method and VDM-EVFI, and the second frame for E2VID. As shown in Tab.~\ref{tab:speed}, while our method exhibits a higher latency compared to the non-diffusion-based E2VID, it significantly outperforms VDM-EVFI, demonstrating superior efficiency among diffusion-based frameworks.

\begin{table}[t]
\centering
\caption{\textbf{Comparison of inference speed.}}
\label{tab:speed}
\begin{tabular}{lc}
\toprule
Method & Inference Speed (s/frame) \\ \midrule
E2VID & 0.0024 \\
VDM-EVFI & 5.4483 \\
Ours & 3.5964 \\ \bottomrule
\end{tabular}
\end{table}

\section{Dataset sizes.}
Our model is trained on the BS-ERGB training set, which comprises 7,636 frames. To evaluate its performance in reconstruction and prediction, we utilize the EVREAL benchmark, encompassing the ECD (1,855 frames), MVSEC (11,321 frames), and HQF (15,499 frames) datasets. For the frame interpolation task, evaluation is conducted on the full HQF dataset (15,513 frames) and the BS-ERGB test set (4,546 frames). Despite being trained on the relatively small-scale BS-ERGB dataset, our model delivers strong results across these diverse benchmarks, demonstrating its robust generalization capabilities.

\end{document}